\documentclass[review]{elsarticle}
\usepackage{lineno,hyperref}
\usepackage{color}
\usepackage{adjustbox}
\usepackage[T1]{fontenc}
\usepackage[utf8]{inputenc}
\usepackage{lineno, hyperref}
\usepackage{graphicx, float}
\usepackage{enumitem}
\usepackage[table,xcdraw]{xcolor}
\usepackage{float}
\usepackage{geometry}
\usepackage{tgadventor}
\usepackage[section]{placeins}

\journal{arXiv preprint}

\biboptions{authoryear}

\begin{document}

\begin{frontmatter}
  \title{Mitigating Domain Drift in Multi Species Segmentation with DINOv2: A Cross-Domain Evaluation in Herbicide Research Trials}



  \author[Tecnalia_address,ehu_address]{Artzai Picon\corref{mycorrespondingauthor}}
  \ead{artzai.picon@tecnalia.com}
  \cortext[mycorrespondingauthor]{Corresponding author at: TECNALIA, Basque Research and Technology Alliance (BRTA), Parque Tecnológico de Bizkaia, C/ Geldo. Edificio 700, E-48160 Derio
    - Bizkaia, Spain}
  \author[Tecnalia_address,ehu_address]{Itziar Eguskiza}
  \ead{itziar.eguskiza@tecnalia.com}
  \author[Tecnalia_address]{Daniel Mugica}
  \ead{daniel.mugica@tecnalia.com}
  \author[BASFESP_address]{Javier Romero}
  \ead{javier.romero@basf.com}
  \author[BASFESP_address]{Carlos Javier Jimenez}
  \ead{carlos-javier.jimenez@basf.com}
  \author[BASFUS_address]{Eric White}
  \ead{eric.louis.white@basf.com}
  \author[BASF_address]{Gabriel Do-Lago-Junqueira}
  \ead{gabriel.junqueira@basf.com}
  \author[BASF_address]{Christian Klukas}
  \ead{christian.klukas@basf.com}

  \author[BASF_address]{Ramon Navarra-Mestre}
  \ead{ramon.navarra-mestre@basf.com}

  \address[Tecnalia_address]{TECNALIA, Basque Research and Technology Alliance (BRTA), Parque Tecnol\'{o}gico de Bizkaia, C/ Geldo. Edificio 700, E-48160 Derio - Bizkaia (Spain)}
  \address[ehu_address]{University of the Basque Country, Plaza Torres Quevedo, 48013 Bilbao (Spain)}
  \address[BASFUS_address]{BASF Corporation, 26 Davis Drive, North Carolina 27709-3528 Research Triangle Park (USA)}
  \address[BASFESP_address]{BASF Espanola S.L. Carretera A376, 41710 Utrera, Sevilla (Spain)}
  \address[BASF_address]{BASF SE, Speyererstrasse 2, 67117 Limburgerhof (Germany)}

  \begin{abstract}
    We evaluate the effect of domain drift on plant species and herbicide‑damage segmentation in realistic, complex agricultural scenarios, and quantify how vision foundation models alter robustness under those shifts. Models trained on controlled, in‑distribution datasets commonly suffer large performance drops when faced with seasonal, geographic, device, and sensor variability; our goal is to measure those drops and assess whether large pretrained backbones mitigate them rather than to introduce a new segmentation method.

    Concretely, we analyze segmentation performance on a large, multi‑year field dataset collected in Germany and Spain (2018–2020) covering 14 plant species and 4 damage classes, and evaluate generalization across three real‑world shift axes: temporal/device drift (2023), geographic transfer to the United States, and extreme sensor/viewpoint shift to drone imagery (2024). We compare a foundation‑model backbone (DINOv2) combined with hierarchical taxonomic inference against prior baselines to isolate the impact of pretrained representations on robustness.

    Our evaluation shows foundation models substantially reduce degradation from domain drift: species‑level F1 increases from 0.52 to 0.87 on in‑distribution data, and foundation backbones retain far more performance under shift (moderate shift: 0.77 vs. 0.24; extreme aerial shift: 0.44 vs. 0.14). Hierarchical taxonomic inference further cushions failures by producing useful coarser‑level predictions when species‑level labels break down (e.g., family F1: 0.68, class F1: 0.88 on aerial imagery). Error analysis reveals that failures under severe shift stem primarily from vegetation–soil confusion, suggesting that taxonomic distinctions remain preserved despite background and viewpoint variability.

    These results provide empirical evidence that foundation visual models meaningfully mitigate domain drift in operational phenotyping pipelines, and they offer practical guidance for deploying robust segmentation systems in herbicide research trials across diverse regions and sensing setups.

  \end{abstract}

  \begin{keyword}
    \begin{footnotesize}
      Large Visual Models \sep Plant Species and Damage segmentation \sep precision agriculture \sep Precise Phenotyping for Herbicide Research Trials \sep Domain shift
    \end{footnotesize}
  \end{keyword}
\end{frontmatter}


\section{Introduction}
\label{ssec:introduction}

The advancement of deep learning and image-based AI has significantly increased awareness of automated weed identification. Publications addressing automated weed identification have increased tenfold since 2010 (\cite{coleman2022weed}). This surge is due to the advent of deep learning technologies, which can better handle the variability of large-scale cropping systems for in-crop weed recognition (\cite{coleman2022weed}).

To manage weed populations and optimize crop yields, farmers rely on herbicides, which necessitates the development of more effective herbicides with greater sensitivity to different species. This process requires extensive R\&D field trials, where parameters such as vigor, crop cover, and species cover are quantified through visual assessments. These assessments depend heavily on the expertise of the field researcher, making them tedious, time-consuming, and susceptible to human variability. Automated weed segmentation algorithms, which serve to unequivocally identify each pixel in a field image, offer a promising solution to address these challenges by providing objective, consistent, and scalable assessments.

Extensive work has been done on detecting and segmenting crops and weeds in images using deep learning approaches. Most recent research (see Table \ref{tab:evaluation_papers}) focuses on the crop/generic weed identification problem (\cite{bi2022development,gao2024cross,ilyas2023overcoming,gupta2023multiclass,you2020dnn, modi2023automated}), as this may be sufficient for automated weed removal systems. Other studies cover more species. For example, \cite{wang2022weed25} generated a dataset of 25 different weed species in the Chongqing region, whereas \cite{zou2023multi} covered six species in two different fields. However, only \cite{gao2024cross, picon_weeds_2021, picon_weeds_2025} supported data acquisition in different fields and regions. Regarding drone data, only \cite{gao2024cross}, \cite{weyler2023pami} and \cite{rai2024weedvision} evaluated weed recognition using drone data. For damage estimation, only \cite{picon_weeds_2025} evaluated the effect and identification of types of damages in crops and weeds.

Despite these advances, two critical limitations prevent existing systems from meeting the needs of real-world herbicide trials. First, analyzing damage across various weed species is crucial for accurate herbicide efficacy assessment, yet most studies do not address damage identification. Second, although the relevance of measuring the resilience of algorithms to domain drifts has been advised (\cite{zhang2023early, rai2023applications}), most studies do not tackle this problem. In herbicide research and development, domain variability represents a fundamental challenge that directly impacts the reliability and cost-effectiveness of field trials. Herbicide efficacy varies significantly across different temporal conditions (growing seasons, weather patterns), geographical locations (soil types, climate zones), and species populations (genetic diversity, resistance patterns).

\begin{table}[h]
  \centering
  \resizebox{\textwidth}{!}{%
    \begin{tabular}{p{0.15\linewidth}p{0.25\linewidth}p{0.07\linewidth}p{0.15\linewidth}p{0.20\linewidth}p{0.15\linewidth}p{0.15\linewidth}p{0.15\linewidth}}
      \hline
                                      & Supported                                  & Supported & Supported                      & Evaluated on     & Evaluated on   & Evaluated on  & Evaluated on \\
      Work                            & Species                                    & Damages   & Locations                      & Unseen Locations & Unseen Sensors & Ulterior Data & Drone data   \\
      \hline
      \cite{wang2022weed25}           & 25                                         & 0         & Chongqing                      & No               & No             & No            & No           \\
      \cite{picon_weeds_2025}         & 14                                         & 4         & multiple fields in 2 countries & No               & No             & No            & No           \\
      \cite{picon_weeds_2021}         & 9                                          & 0         & multiple fields in 2 countries & No               & No             & No            & No           \\
      \cite{modi2023automated}        & 2 (crop, generic weed)                     & 0         & 1 location                     & No               & No             & No            & No           \\
      \cite{calderara2024two}         & 3                                          & 0         & 1 location                     & No               & No             & No            & No           \\
      \cite{li2024winter}             & 8                                          & 0         & 1 field                        & No               & No             & No            & No           \\
      \cite{zou2023multi}             & 6                                          & 0         & 2 fields                       & No               & No             & No            & No           \\
      \cite{kamath2022classification} & 3 (crop, broad-leaf weed, grass-leaf-weed) & 0         & 1 location                     & No               & No             & No            & No           \\
      \cite{bi2022development}        & 2 (crop, generic weed)                     & 0         & 1 field                        & No               & No             & No            & No           \\
      \cite{gao2024cross}             & 2 (crop, generic weed)                     & 0         & four fields in 2 regions       & Yes              & Yes            & No            & Yes          \\
      \cite{ilyas2023overcoming}      & 2 (crop, generic weed)                     & 0         & 4 fields                       & Yes              & No             & No            & No           \\
      \cite{gupta2023multiclass}      & 2 (crop, generic weed)                     & 0         & 1 farm                         & No               & No             & No            & No           \\
      \cite{you2020dnn}               & 2 (crop, generic weed)                     & 0         & 1 location                     & No               & No             & No            & No           \\
      \cite{weyler2023pami}           & 2 (crop, generic weed)                     & 0         & 2 locations                    & No               & No             & Yes           & Yes          \\
      \cite{rai2024weedvision}        & 3 weed species                             & 0         & multiple locations             & No               & No             & No            & yes          \\
      \hline
    \end{tabular}
  }
  \caption{Works targeting species and damage identification and their validation procedure.}
  \label{tab:evaluation_papers}
\end{table}

As shown in Table \ref{tab:evaluation_papers}, most studies do not evaluate their performance on unseen locations or different sensors, which limits their generalizability to novel environments and conditions. This represents a critical gap in the literature and precludes the use of these systems in real-world applications, as domain variability across years, devices, and species constitutes one of the most pressing challenges for in-field AI systems. Unlike controlled laboratory settings, agricultural environments exhibit substantial temporal drift due to seasonal variations, climate change effects, and evolving farming practices. Additionally, the proliferation of different imaging devices—from smartphone cameras to specialized agricultural sensors and drones—introduces significant technical variability in image characteristics such as color profiles, resolution, and noise patterns. Most importantly, the emergence of new weed species and the adaptation of existing species to herbicide treatments creates a continuously evolving biological challenge that existing models fail to address.

Prior work has predominantly evaluated models under in-domain conditions, where training and testing data share similar temporal, geographical, and technical characteristics (\cite{bi2022development,gao2024cross,ilyas2023overcoming,gupta2023multiclass,you2020dnn,picon_weeds_2021}). This evaluation paradigm severely overestimates model performance and fails to capture the reality of deployed agricultural systems, where models must operate effectively across multiple growing seasons, different geographical regions, and varying acquisition conditions. To address these fundamental challenges of domain shift and generalization in agricultural computer vision, we advance two key methodological innovations: (1) pre-training with DINOv2 ({\cite{oquab2023dinov2}}), a self-supervised vision transformers, and (2) hierarchical inference for taxonomical classification that allows us to fall back to broader categories when fine-grained predictions are uncertain. We hypothesize that DINOv2's self-supervised pre-training on diverse natural images will provide more robust visual representations that can better generalize across different agricultural domains compared to traditional CNN architectures.

To test this hypothesis, we designed a comprehensive three-stage evaluation strategy that progressively increases the challenge of domain shift. First, we trained our model on a multi-year dataset collected in Germany, Spain, and the United States between 2018 and 2020 using digital and mobile-phone cameras, establishing baseline performance under controlled conditions. Second, we tested the model on a reality check dataset collected in 2023 from Germany, Spain, and the United States, measuring performance degradation under temporal drift and new species. Finally, we assessed the model's effectiveness using a novel dataset collected in 2024 with drone images, testing generalization to an entirely different imaging modality. This systematic evaluation across temporal, geographical, and device domains provides a more realistic assessment of model robustness and enables us to quantify the benefits of our DINOv2-based approach over traditional CNN baselines in handling real-world agricultural variability.

\section{Datasets}
\label{ssec:datasets}

In this section, we describe three complementary datasets that were generated to develop and validate our approach. These datasets were collected during multiple field campaigns conducted in Germany, Spain, and the United States between 2018 and 2024. The first dataset, BASE, was used for initial model development and training (2018-2020). To assess model generalization to new acquisition conditions, we created the REALITY dataset (2023) using mobile devices and digital cameras. Finally, to evaluate robustness under significant domain shift, we collected the drone-based REALITY dataset (2024) using aerial imagery at ultra-low altitude. Together, these datasets provide a comprehensive benchmark for evaluating weed detection algorithms across varying imaging platforms, environmental conditions, and geographical locations.

\subsection{BASE dataset (2018--2020)}
\label{ssec:train_datasets}

The BASE dataset was developed as the foundation for model training and initial validation. This dataset, previously described in \cite{picon_weeds_2025}, consists of manually annotated images collected from real field plots between 2018 and 2020 across multiple locations. The dataset captures 14~different crop and weed species under various growth conditions and damage states.

The species included are:

\textit{Abutilon theophrasti} (eppo: abuth),
\textit{Amaranthus retroflexus} (eppo: amare),
\textit{Chenopodium album} (eppo: cheal),
\textit{Digitaria sanguinalis} (eppo: digsa),
\textit{Echinochloa crus-galli} (eppo: echcg),
\textit{Portulaca oleracea} (eppo: porol),
\textit{Setaria viridis} (eppo: setve),
Maize, \textit{Zea mays} (eppo: zeamx),
\textit{Datura stramonium} (eppo: datst),
\textit{Echinochloa colona} (eppo: echco),
\textit{Glycine max} (eppo: glxma),
\textit{Helianthus annuus} (eppo: helan),
\textit{Polygonum convolvulus} (eppo: polco), and
\textit{Solanum nigrum} (eppo: solni).

Each species is identified according to the European and Mediterranean Plant Protection Organization (EPPO) code (\cite{eppo1994guideline}, \cite{ayllon2023eppo}). The dataset systematically documents various types of herbicide-induced damage including Initial, Initial-bleaching, Initial-Necrosis, Bleaching, Necrosis, and Leaf-Curling, along with annotated growth stages.

The BASE dataset comprises two distinct collection strategies: Type~A datasets contain real field images with naturally occurring mixtures of multiple species, while Type~C datasets contain single-species images where other species were manually removed during image acquisition. This diversity in collection strategies ensures representation of both monoculture and polyculture field conditions. Images were collected from different locations and environmental conditions, resulting in natural variability in species abundance and distribution across field plots.

\begin{figure*}[ht!]
  \centering
  \small
  \includegraphics[width=12.5cm]{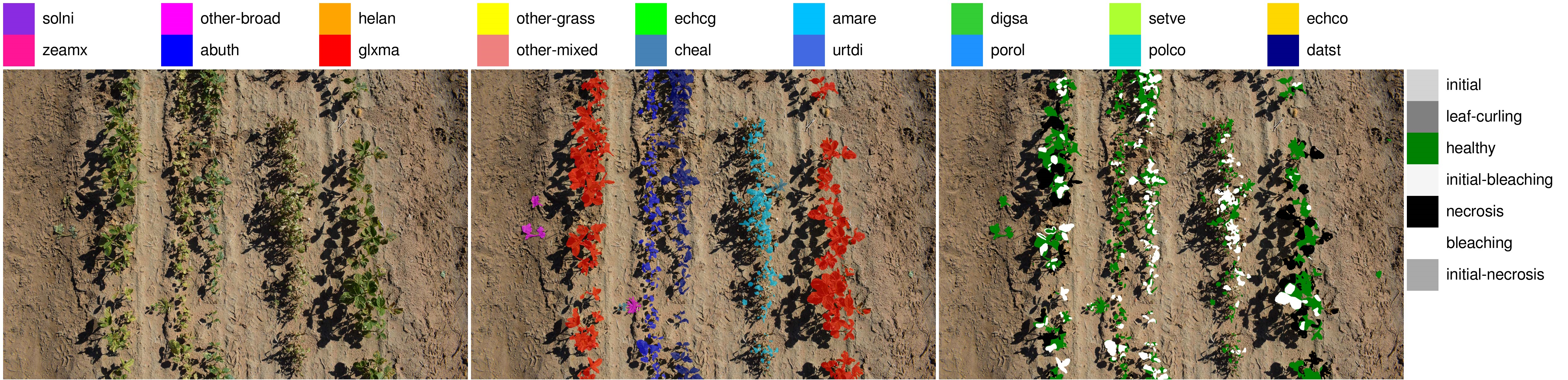}
  \caption{2019A2 dataset. Example image with species and damage masks. Left) Original Image, Middle) Species annotation, Right) Damage annotation.}
  \label{fig:2019A2_fig4_1}
\end{figure*}

All images were systematically annotated using CVAT (\cite{cvat}) and refined with existing vegetation segmentation models. Metadata including dataset location and image field of view were also annotated to enable scale correction and data augmentation. Cases involving annotation uncertainty or impossibility were labeled using specific class identifiers: other-broad (unknown broadleaf species), other-grass (unknown grassleaf species), or other-mixed (mixed species). Table~\ref{tab:real_field_datasets} summarizes the content of the BASE dataset, and Figure~\ref{fig:2019A2_fig4_1} shows an example of annotated species and damage masks.

\begin{table}[ht!]
  \centering
  \footnotesize
  \resizebox{\textwidth}{!}{%
    \begin{tabular}{p{0.08\linewidth}p{0.04\linewidth}p{0.06\linewidth}p{0.34\linewidth}p{0.20\linewidth}p{0.15\linewidth}}
      \hline
      Collection & Type & Images & Species                                                                                          & Damage                                                            & Location       \\
      \hline
      2018A1     & A    & 1107   & abuth, amare, cheal, digsa, echcg, porol, setve, zeamx                                           & None                                                              & Spain          \\
      2019A1     & A    & 147    & abuth, amare, cheal, datst, digsa, echcg, echco, glxma, helan, polco, porol, setve, solni, zeamx & Initial, Initial-bleaching, Initial-Necrosis, Bleaching, Necrosis & Spain, Germany \\
      2019A2     & A    & 282    & abuth, amare, cheal, datst, digsa, echcg, echco, glxma, helan, polco, porol, setve, solni, zeamx & Initial-bleaching, Initial-Necrosis, Bleaching, Necrosis          & Spain, Germany \\
      2019C1     & C    & 249    & abuth, amare, cheal, datst, digsa, echcg, glxma, helan, polco, porol, setve, solni, zeamx        & None                                                              & Spain          \\
      2019C2     & C    & 303    & abuth, amare, cheal, datst, digsa, echcg, glxma, helan, polco, porol, setve, solni, zeamx        & Necrosis                                                          & Spain          \\
      2020CDE1   & C    & 139    & abuth, amare, cheal, digsa, echcg, polco, porol                                                  & Necrosis                                                          & Germany        \\
      2020CES1   & C    & 222    & abuth, amare, cheal, datst, digsa, echcg, echco, glxma, helan, polco, porol, setve, solni        & Bleaching, Necrosis, Leaf-Curling                                 & Spain          \\
      \hline
    \end{tabular}
  }
  \caption{BASE dataset content summary}
  \label{tab:real_field_datasets}
\end{table}

An example of annotated images from the BASE dataset is provided in Figure~\ref{fig:2019A2_fig4_1}, illustrating the annotation of both species and damage masks.

\subsection{REALITY dataset (2023)}
\label{ssec:reality_check_datasets}

To evaluate model robustness and generalization capabilities under different acquisition conditions, we created the REALITY dataset in 2023. This dataset was specifically designed to assess algorithm performance on previously unseen imaging equipment, locations, and environmental conditions that were not represented in the training data.

Images were acquired in three countries (Germany, Spain, and the United States) using consumer-grade equipment: a BASF mobile application running on a Samsung A54 smartphone and a Nikon D750 digital camera. This contrasts with the BASE dataset's acquisition setup, introducing significant domain shift to test model robustness. All images were manually annotated using CVAT(\cite{cvat}) following the same protocol as the BASE dataset.

\begin{figure*}[ht!]
  \centering
  \small
  \includegraphics[width=12cm]{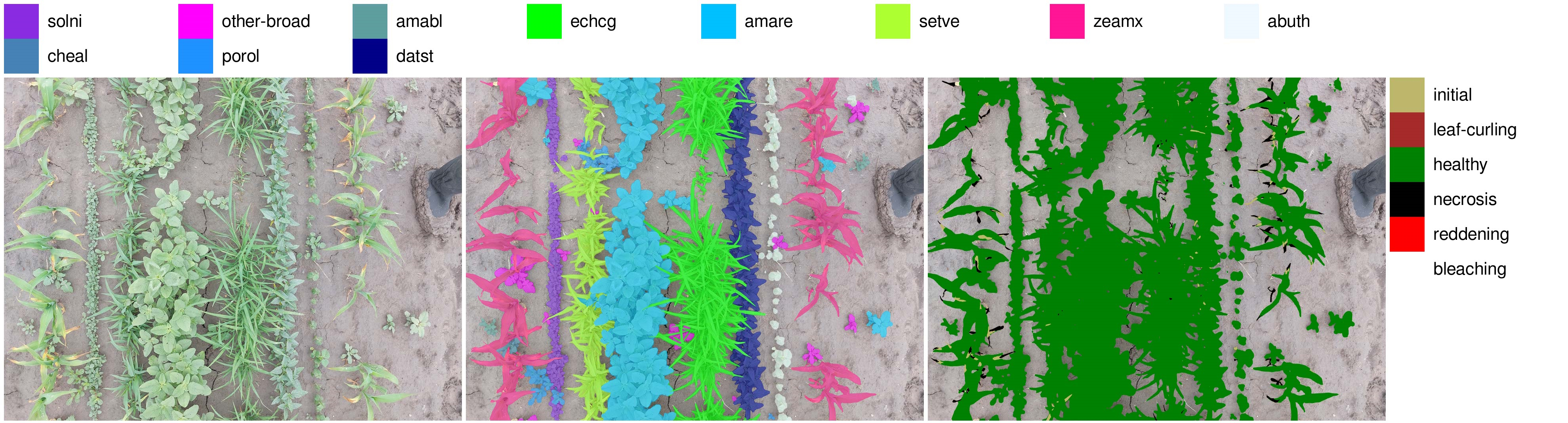}
  \includegraphics[width=12cm]{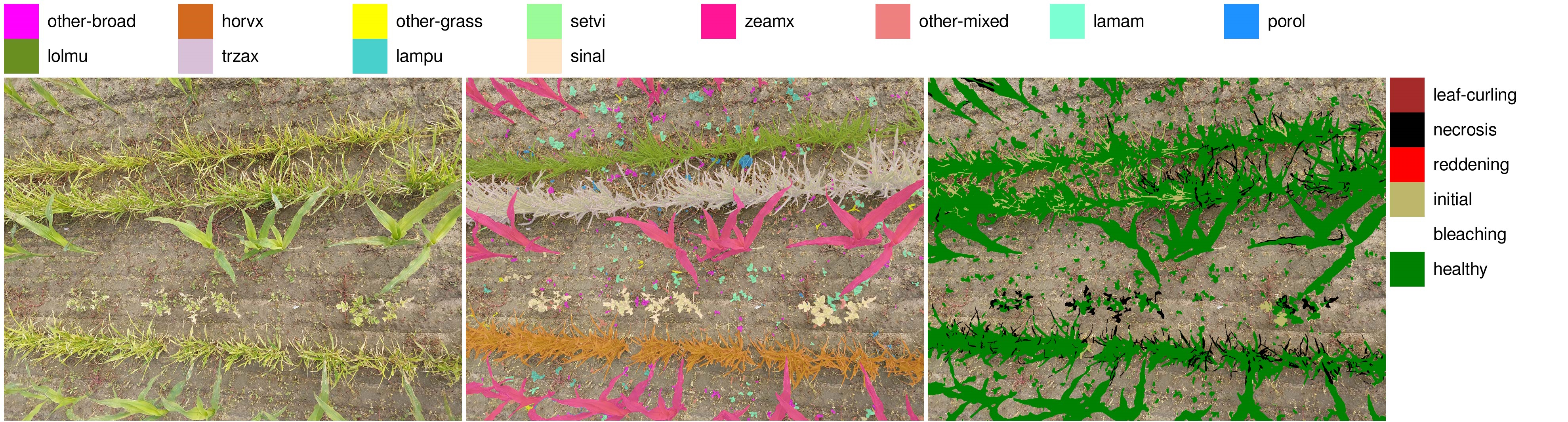}
  \includegraphics[width=12cm]{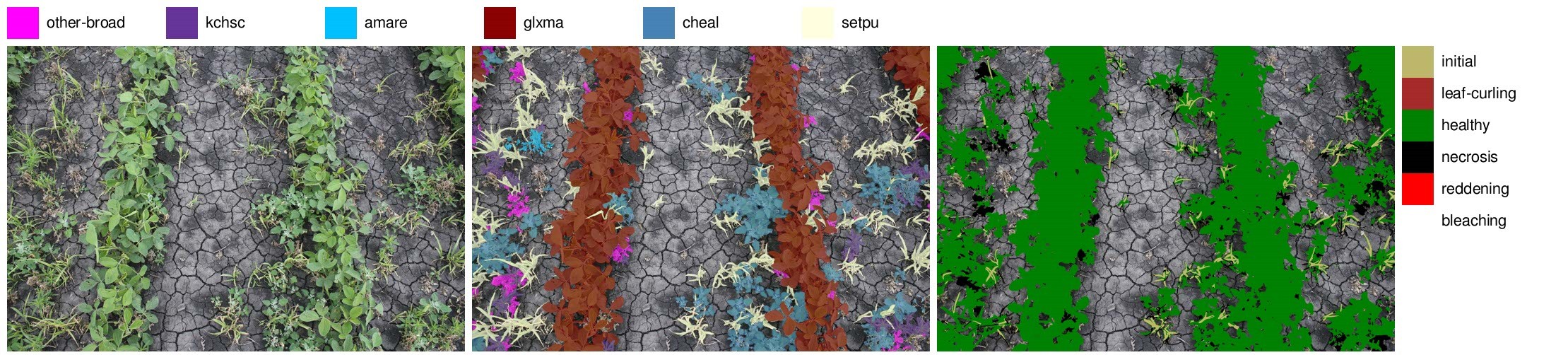}
  \caption{Reality check dataset examples. Top) Germany, Middle) Spain, and Bottom) United States. Each figure shows an image with species and damage masks. Left) Original Image, Middle) Species annotation, Right) Damage annotation.}
  \label{fig:dataset_reality_example}
\end{figure*}

The REALITY dataset encompasses 35~different crop and weed species, significantly expanding the taxonomic coverage beyond the BASE dataset. The species include: \textit{Amaranthus hybridus} (amach), \textit{Chenopodium album} (cheal), \textit{Datura stramonium} (datst), \textit{Echinochloa crus-galli} (echcg), \textit{Glycine max} (glxma), \textit{Helianthus annuus} (helan), \textit{Hordeum vulgare} (horvx), \textit{Lamium amplexicaule} (lamam), \textit{Lamium purpureum} (lampu), \textit{Lolium multiflorum} (lolmu), \textit{Mercurialis annua} (meran), \textit{Polygonum aviculare} (polam), \textit{Polygonum convolvulus} (polco), \textit{Polygonum persicaria} (polpe), \textit{Portulaca oleracea} (porol), \textit{Setaria viridis} (setve), \textit{Setaria verticillata} (setvi), \textit{Sinapis alba} (sinal), \textit{Spergula arvensis} (sprar), \textit{Stellaria media} (steme), \textit{Triticum aestivum} (trzax), \textit{Zea mays} (zeamx), \textit{Amaranthus blitoides} (amabl), \textit{Amaranthus retroflexus} (amare), \textit{Digitaria sanguinalis} (digsa), \textit{Solanum nigrum} (solni), \textit{Amaranthus palmeri} (amapa), \textit{Amaranthus tuberculatus} (amatu), \textit{Conyza canadensis} (conar), \textit{Cyperus esculentus} (cypes), \textit{Eleusine indica} (elein), \textit{Kochia scoparia} (kchsc), \textit{Mollugo verticillata} (molve), \textit{Polygonum pensylvanicum} (polpy), and \textit{Setaria pumila} (setpu). As with the BASE dataset, uncertain identifications were annotated using the other-broad, other-grass, or other-mixed classes. Table~\ref{tab:reality_dataset} summarizes the dataset content, and Figure~\ref{fig:dataset_reality_example} presents representative examples from each country.

\begin{table}[ht!]
  \centering
  \footnotesize
  \resizebox{\textwidth}{!}{%
    \begin{tabular}{p{0.08\linewidth}p{0.04\linewidth}p{0.06\linewidth}p{0.34\linewidth}p{0.20\linewidth}p{0.15\linewidth}}
      \hline
      Collection & Type & Images & Species                                                                                                                                                   & Damage                                     & Location      \\
      \hline
      2023ADE    & A    & 32     & amach, cheal, datst, echcg, glxma, helan, horvx, lamam, lampu, lolmu, meran, polam, polco, polpe, porol, setve, setvi, sinal, sprar, steme, trzax, zeamx, & Initial, Bleaching, Necrosis, Leaf-Curling & Germany       \\

      2023AES    & A    & 36     & amabl, amare,  cheal, datst, digsa, echcg, glxma, helan, polco, porol, setve, setvi, solni, zeamx,                                                        & Initial, Necrosis, Leaf-Curling            & Spain         \\

      2023AUS    & A    & 37     & amabl, amapa,  amare, amatu, cheal, conar, cypes, digsa, echcg, elein, glxma, helan, kchsc, molve, polpy, porol, setpu, setvi, zeamx,                     & Initial, Bleaching, Necrosis, Leaf-Curling & United States \\

      \hline
    \end{tabular}
  }
  \caption{REALITY dataset content summary}
  \label{tab:reality_dataset}
\end{table}

\subsection{Drone-based REALITY dataset (2024)}
\label{ssec:reality_check_drone_datasets}

To assess algorithm performance under substantial domain shift, we collected a drone-based dataset in late 2023 and throughout 2024. This dataset represents the most challenging evaluation scenario, introducing significant changes in imaging perspective (aerial vs.\ ground-level), spatial resolution, and field coverage compared to the BASE and REALITY datasets.

Images were acquired using two DJI drone platforms: Matrice 300 RTK and Matrice 350 RTK, both equipped with DJI Zenmuse P1 cameras. Flights were conducted at 6~meters above ground level across field plots in Spain, Germany, and the United States. Two focal lengths were employed: 35~mm (used at all locations, providing 0.75~mm/pixel ground sampling distance) and 50~mm (used in the USA only, providing 0.52~mm/pixel ground sampling distance). The flight paths were planned based on plot dimensions, and images were subsequently cropped according to plot shapefiles to ensure accurate spatial alignment.

The dataset captures four crop species: maize (\textit{Zea mays}, zeamx), sunflower (\textit{Helianthus annuus}, helan), soybean (\textit{Glycine max}, glxma), and cotton (\textit{Gossypium hirsutum}, goshi). Cotton was included in the dataset but not in the evaluation as it represents a new crop still undergoing annotation. The complete dataset encompasses a significantly broader range of weed species compared to previous collections, with annotations for over 70~species across all three geographical regions.

Table~\ref{tab:sensor_specs} provides detailed specifications for the imaging sensors and their footprint characteristics at the 6-meter flight altitude.

\begin{table}[ht!]
  \centering
  \footnotesize
  \caption{Specifications of the DJI Zenmuse P1 Sensors and their approximate imaging footprint at 6 meters of height.}
  \label{tab:sensor_specs}
  \begin{tabular}{|c|c|c|c|c|c|}
    \hline
    \textbf{Sensor} & \textbf{Height (m)} & \textbf{Focal length (mm)} & \textbf{GSD (mm/pixel)} & \textbf{Width/Height (px)} & \textbf{Width/Height (m)} \\
    Zenmuse P1      & 6                   & 35                         & 0.7543                  & 8192x5460                  & 6.18x4.12                 \\
    \hline
    Zenmuse P1      & 6                   & 50                         & 0.5280                  & 8192x5460                  & 4.33x2.88                 \\
    \hline
  \end{tabular}
\end{table}

\begin{figure}[ht!]
  \centering
  \footnotesize
  \caption{Drone-based reality dataset content summary}
  \label{tab:reality_dataset_drones}
  \includegraphics[width=1\linewidth]{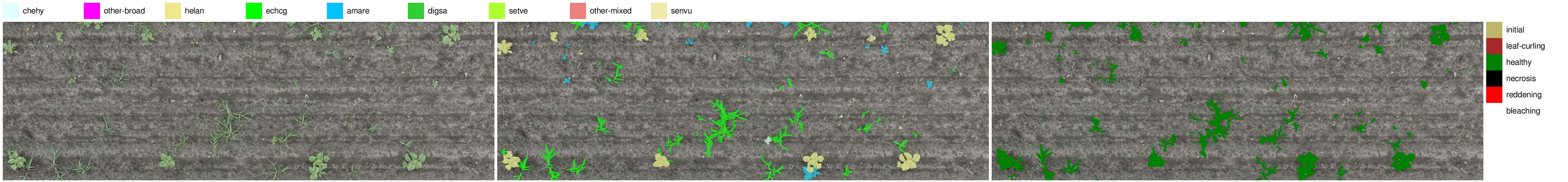}
  \caption{Drone-based reality dataset. Example image with species and damage masks. Left) Original Image, Middle) Species annotation, Right) Damage annotation.}
  \label{fig:example_dataset_reality_uav}
\end{figure}

An example image is shown in Figure~\ref{fig:example_dataset_reality_uav}. As it can be seen the field of view of the drone images is much larger than the images on the BASE and REALITY datasets. Image color appearance is also different. A summary of the dataset can be found in Table~\ref{tab:reality_dataset_drones}.

\begin{table}[ht!]
  \centering
  \footnotesize
  \resizebox{\textwidth}{!}{%
    \begin{tabular}{p{0.08\linewidth}p{0.04\linewidth}p{0.06\linewidth}p{0.34\linewidth}p{0.20\linewidth}p{0.15\linewidth}}
      \hline
      Collection & Type & Images & Species                                                                                                                                                                                                                                                                                                                                                                                                                                                                                                                                                             & Damage                                     & Location                      \\
      \hline
      UAV241122  & A    & 201    & abuth, accos, alomy, amabl, amach, amapa, amare, amata, amatu, angar, avefa, brsnn, capbp, casob, cheal, chehy, cirar, conar, cypcp, cypes, datst, digsa, dttae, dttss, ecael, echcg, echco, elein, galap, glxma, goshi, helan, hisin, horvx, ipohe, ipola, kchsc, lamam, lampu, lolmu, meran, molve, other-broad, other-grass, other-mixed, pandi, panmi, pesgl, pibsa, polam, polco, polpe, polpy, porol, porpi, rapra, rchsc, senvu, setpu, setve, setvi, sidsp, sinal, soil, solni, soltu, sonol, sorvu, sprar, steme, trbte, trzax, urtdi, urtur, vioar, zeamx & Initial, Bleaching, Necrosis, Leaf-Curling & Germany, Spain, United States \\
      \hline
    \end{tabular}
  }

\end{table}

\subsection{Analysis of drift}

\begin{figure*}[ht!]
  \centering
  \small
  \includegraphics[width=10.5cm]{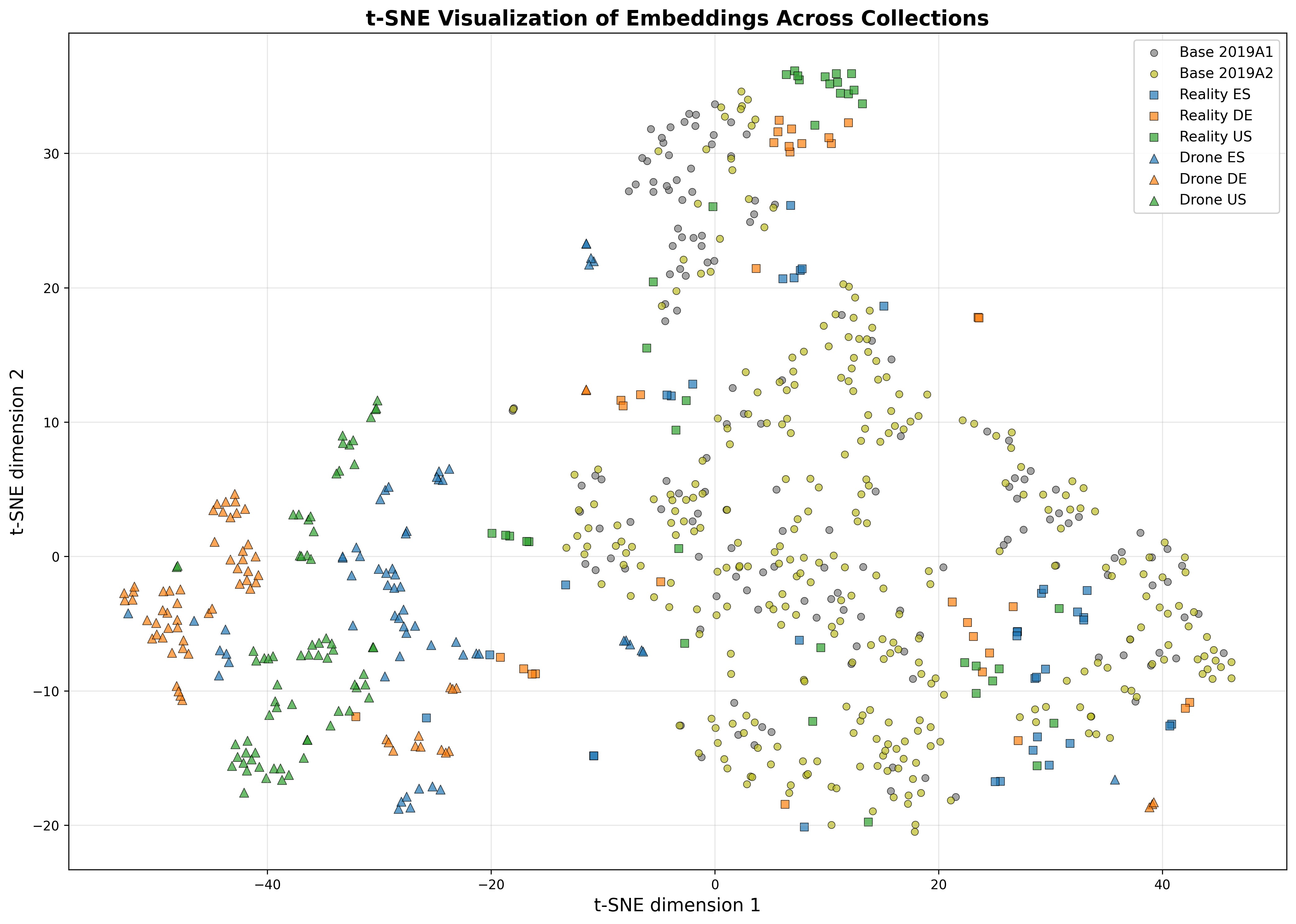}
  \caption{t-SNE visualization of all dataset collections, illustrating the distribution of embeddings extracted from a DINOv2 model from images across the different datasets. Each point represents a different image, and each color represents a different location. The datasets used for training (Base) and the REALITY dataset and DRONE dataset are shown with different geometrical shapes.}
  \label{fig:tsne_all_collections}
\end{figure*}

The analysis of drift in the datasets reveals that datasets without drift, such as 2018A1 and 2019A2, which were used for training and testing without any drift, show the lowest drift. On the other hand, the reality check dataset taken with mobile phones and digital cameras shows higher drift, as can be seen in Figure~\ref{fig:tsne_all_collections}. The highest drift comes from the drone images dataset, which was the most challenging due to the different acquisition method and the larger variety of locations and environmental conditions. Embeddings were extracted using a DINOv2 model (\cite{oquab2023dinov2}) and visualized using t-SNE (\cite{maaten2008visualizing}).

\section{Proposed Method}
\subsection{Large Visual Model (DinoV2) Based Architecture}
\label{ssec:architecture}
Previous work (\cite{picon_weeds_2025}) was based on an EfficientNet architecture (\cite{tan2019EfficientNet}) with a DeepLabV3$+$ decoder (\cite{deeplab}). EfficientNet is known for its ability to achieve high performance with fewer parameters and lower computational cost.

On top of this backbone, three independent decoders were used to address the tasks of vegetation segmentation, species identification, and damage classification. Each decoder was designed to specialize in its respective task while sharing a common feature extraction backbone. This multi-decoder approach enabled the model to leverage shared representations from the EfficientNet backbone while tailoring the final predictions to the specific requirements of each task.

Although independent decoders are not strictly necessary, previous work has shown that this approach can enhance performance in multi-task learning scenarios by allowing each decoder to focus on the unique aspects of its task while also helping the backbone learn more general features (\cite{yu2024unleashing}, \cite{picon2025crop}). Additionally, computational efficiency is improved by sharing the backbone across tasks, reducing overall model size, training time, and inference cost compared to using separate models for each task.

In this work, we have improved this network by integrating a more modern self-supervised large visual model based on the DinoV2 architecture (\cite{oquab2023dinov2}). The selected decoder is based on SegFormer's multi-scale decoder (\cite{xie2021segformer}) to seamlessly incorporate information from different scales. To address the different tasks, we used a common backbone for the three tasks with three independent decoders. The last layer of each decoder was removed, and a final convolutional layer of spatial size (1,1) was added per task. The number of filters in this layer corresponded to the number of classes in the task. The activation function for the species and damage tasks was softmax.

This architecture allows the use of independent decoder networks while employing a single encoder for all tasks.
The loss functions for the vegetation, species, and damage tasks were weighted categorical cross-entropy, where the weights were extracted using the effective number of samples method (\cite{cui2019class}), extended to semantic segmentation tasks. We employed an unbalancing beta parameter of 0.99.

\begin{figure*}[ht!]
  \centering
  \small
  \includegraphics[width=12.5cm]{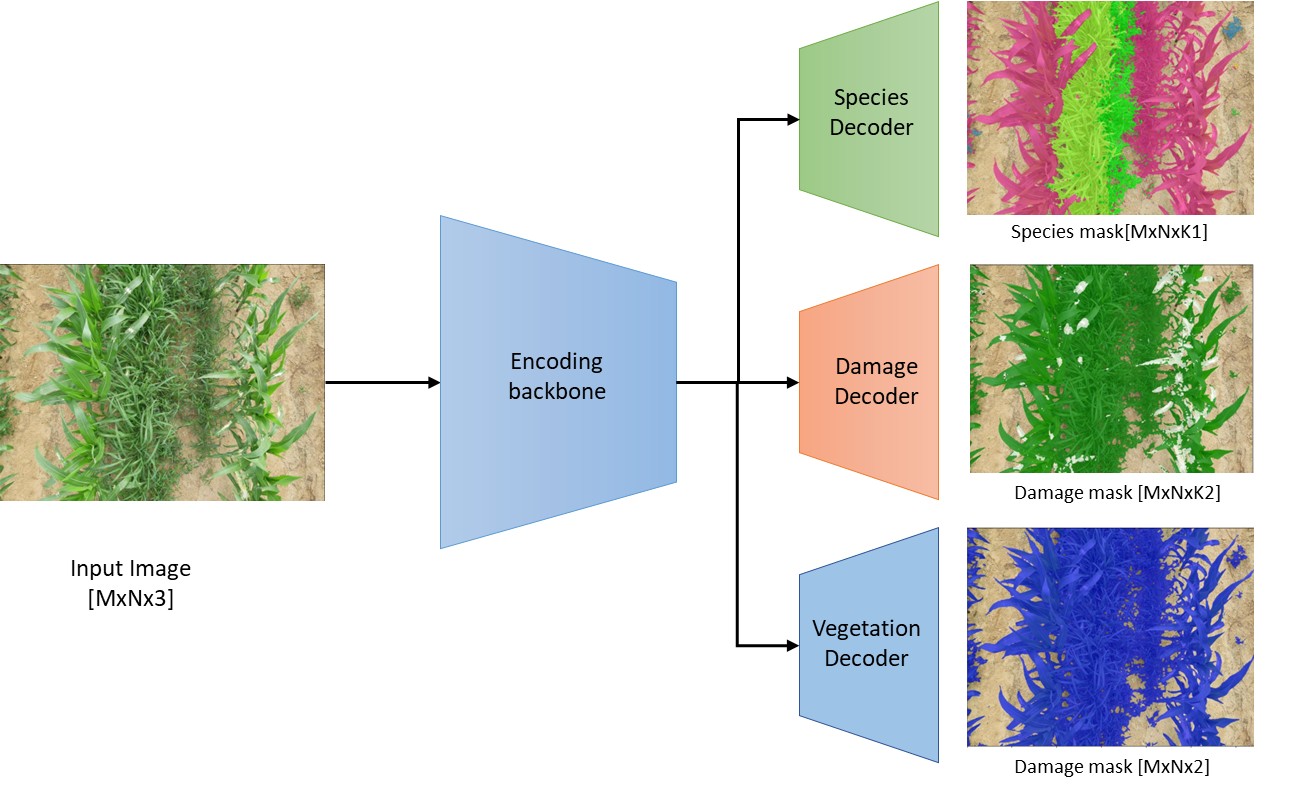}
  \caption{Proposed architecture. A common encoding backbone is connected to the three independent decoders. Each decoder handles one of the network tasks (vegetation, damage, and species). K1 represents the number of species classes, and K2 represents the number of damage classes.}
  \label{fig:diagram_architectures}
\end{figure*}

\subsection{Hierarchical Inference}
\label{ssec:hierarchical_inference}

During inference, hierarchical inference was implemented. This process is performed in a taxonomical manner for each pixel, similar to the approach described in \cite{picon_weeds_2025}, but applied during inference rather than training. Probabilities are aggregated across different taxonomy levels, similar to the method used by \cite{hoang2023adaptive} for classification tasks, but extended to the pixel level as described in \cite{picon_weeds_2025}.

Hierarchical inference involves organizing the classification tasks into a tree-like structure, where each node represents a different level of the taxonomy. For species recognition tasks, the taxonomies include species, genus, family, order, and class. For damage classification, we established an initial damage-healthy hierarchy before identifying the specific type of damage. During inference, probabilities are aggregated toward their upper class nodes, starting from general nodes and moving toward more specific ones. An argmax is calculated for each hierarchical node, with recursive argmax values generated for the more specific nodes. This method prevents the "spreading" of classification probabilities over slightly different species that share a common genus or family. Additionally, hierarchical confidence is calculated at each node as the probability of the winning node.

The hierarchical inference formula is as defined in equation \ref{eq:hierarchical_inference}:

\begin{equation}
  P_{node_{i+1}} = \sum_{j \in children(node_i)} P_{node_j}
  \label{eq:hierarchical_inference}
\end{equation}

where $P_{node_i}$ is the probability of each class at node $i$ in the hierarchical tree, and $children(node_i)$ is the set of child nodes of node $i$. The probability of the winning class at the parent node is calculated as the sum of the probabilities of its child nodes. For species recognition tasks, the hierarchical structure is as follows: Species $\rightarrow$ Genus $\rightarrow$ Family $\rightarrow$ Order $\rightarrow$ Class $\rightarrow$ Vegetation/background. The model is able to report classifications at any of these taxonomical levels.

For damage classification, the hierarchical structure is as follows: Specific Damage Type $\rightarrow$ Damage/healthy $\rightarrow$ Vegetation/background. The model first identifies whether the image contains damage, and then identifies the specific type of damage. The hierarchical inference mechanism allows the model to report the presence of damage without necessarily identifying the specific damage type. This is particularly useful when the model is uncertain about the specific damage type, as it can still report the presence of damage.

Hierarchical inference allows us to report classifications at different taxonomical levels. In this sense, a single model can provide predictions at varying levels of specificity, from broad categories (e.g., family level) to fine-grained classifications (e.g., species level). In the future, confidence information can be used to report broader classifications when necessary. When the confidence at a specific level is below a predefined threshold, the model can fall back to a higher-level classification. This approach enhances the robustness of the model by providing more reliable predictions, especially in cases where fine-grained classification may be uncertain.

\subsection{Training Procedure}
\label{ssec:training}
For training the proposed architecture, we used the BASE dataset described in Section \ref{ssec:datasets}. This dataset contains images from multiple years (2018-2020), locations (Germany, Spain), and acquisition devices (digital cameras, mobile phones). This diversity in the training data helps the model learn robust features that can generalize well across different conditions. Training, validation, and test subsets were created as described in \cite{picon_weeds_2025}. This selection was performed plot-field-wise to avoid bias. This means that all images from a specific plot-field were assigned to the same subset (train, validation, or test) across time, as several image sequences were acquired from the same plot-field during the acquisition campaigns.

To ensure replicability, the experiments were trained using the train, validation, and test subsets of the BASE dataset, with exactly the same subsets and training procedure as used in \cite{picon_weeds_2025}. Since our datasets have different numbers of images, we implemented a generator that fetches images from the different datasets in an equal manner. It takes one image from each dataset at a time. If a dataset runs out of images, it starts over while maintaining the count for the others. During training, images were randomly cropped to tiles of size 840x840 pixels. This size was selected to balance the need for sufficient context in the images while managing computational resources effectively. During inference tiling was used to process the full-size images with an overlap of 240 pixels between tiles to avoid edge artifacts.

Data augmentation was applied every time a new image was fetched by the data generator. The transformations we applied for data augmentation are: rotation, height and width shift, zoom, vertical and horizontal flip, pixel-intensity shift (color change), Gaussian blur, perspective and artificial shadow simulation. These augmentations were applied using the Albumentations library (\cite{buslaev2020albumentations}).

The model was trained using the AdamW optimizer (\cite{loshchilov2017decoupled}) with a learning rate of $5 \times 10^{-5}$. AdamW was selected for its superior performance in transformer-based architectures due to its decoupled weight decay mechanism, which provides better regularization compared to standard Adam optimization. The relatively low learning rate was chosen to ensure stable convergence during fine-tuning of the pre-trained DINOv2 backbone, preventing catastrophic forgetting of the learned representations while allowing gradual adaptation to the agricultural domain. The momentum parameter of $0.9$ was employed to accelerate convergence and reduce oscillations in the gradient descent process, which is particularly beneficial for the complex multi-task learning scenario involving species classification, damage assessment, and vegetation segmentation tasks.

For the EfficientNet decoder, the encoder weights were frozen for the first 25 epochs, and then the encoder weights were unfrozen for the remaining epochs. For the DinoV2 decoder, the weights were completely frozen during the training process. A total of 200 epochs were selected for training in order to fairly compare among the different experiments.

For training the networks designed in our experiments, we used a NVIDIA Tesla H100 with 80GB of memory. Considering the size of the input images, we set the \textit{batch size} to 12.

\subsection{Evaluation Methodology}
\label{ssec:evaluation}

For evaluating the model, we used both the test subset of the BASE dataset and two independent datasets: REALITY and DRONE, as described in Section \ref{ssec:datasets}. The BASE dataset test subset allows us to assess model performance under no domain shift conditions, while the REALITY dataset (2023) and DRONE dataset (2024) provide a realistic assessment of the model's performance under temporal, geographical, and device domain shifts. These evaluation datasets contain images from different years, locations (Germany, Spain, United States), and acquisition devices (digital cameras, mobile phones, drones), and were not used during training.


During the manual annotation process, some classes were used to facilitate the annotation process for field technicians. These classes include "other-broad", "other-grass", and "other-mixed", which group together various weed species that are not specifically targeted in the study or in situations of difficult annotation due to uncertainty (e.g., jungle images) or to speed up the annotation process. However, these classes do not provide relevant information for evaluating the model's performance in identifying specific weed species or damage types. Therefore, for the purpose of model evaluation, these classes were excluded from the metrics calculation.


To assess the uncertainty in the segmentation performance metrics, we employed an image-level bootstrap resampling approach. For each class, confusion matrix values (true positives, true negatives, false positives, and false negatives) were collected from all images in the dataset. Bootstrap samples (n=1000) were generated by randomly sampling images with replacement, maintaining the original sample size. For each bootstrap iteration, confusion matrix values from the sampled images were aggregated, and performance metrics (sensitivity, specificity, accuracy, positive predictive value, negative predictive value, balanced accuracy, Dice coefficient, and F1 score) were calculated from these aggregated values. Confidence intervals (95\%) were derived using the percentile method, with the lower and upper bounds corresponding to the 2.5th and 97.5th percentiles of the bootstrap distribution, respectively. Classes without positive samples (TP + FN = 0) across all images were excluded from the analysis. Additionally, to evaluate the relationship between annotated and detected class proportions, we calculated the coefficient of determination ($R^2$) across images for each class, with bootstrap confidence intervals estimated using the same resampling strategy. This methodology provides robust estimates of classification performance and uncertainty, accounting for variability across images.

When reporting results at different taxonomical levels using the hierarchical inference strategy described in Section \ref{ssec:hierarchical_inference}, bootstrap confidence intervals were calculated for each taxonomical level using the same resampling strategy, ensuring robust estimates of performance metrics across different levels of the taxonomy.

\section{Results}
\label{sec:results}

\subsection{Evaluation on the BASE Dataset: No Domain Shift Conditions}
\label{ssec:results_test}

The objective of this experiment is to compare the performance of the proposed model with the previous state-of-the-art model presented in \cite{picon_weeds_2025} under no domain shift conditions. To this end, the proposed model was trained and tested with the BASE dataset (see Section~\ref{ssec:train_datasets}) using the same training, validation, and test splits as described in \cite{picon_weeds_2025}.

The results presented are based on the test portion of the 2019A2 dataset. Since we completely updated the development stack, we retrained the model from \cite{picon_weeds_2025} for fair comparison. The use of the self-supervised DINOv2 encoder has led to significantly improved performance across all tasks, including species identification, damage assessment, and vegetation classification. The study in \cite{picon_weeds_2025} already provides an exhaustive analysis of various architectures.

Table~\ref{tab:test_camera_species_experiments_comparison} presents a comprehensive comparison of the proposed DINOv2-based model against the previous DeepLabV3-based model across different taxonomical hierarchies for species identification. The DINOv2 model consistently outperforms the DeepLabV3 model at all hierarchical levels, from species to vegetation. Notably, at the species level, the DINOv2 model achieves a balanced accuracy of $0.935 \pm 0.014$, a precision of $0.870 \pm 0.032$, a recall of $0.874 \pm 0.027$, an F1 score of $0.870 \pm 0.027$, and an $R^2$ of $0.983 \pm 0.017$. In contrast, the DeepLabV3 model records significantly lower metrics in all categories.

Metrics are weighted averages across all classes, where each class is weighted by its total number of samples. Bootstrap analysis was performed at the file level by resampling files with replacement, aggregating confusion matrix values, and computing metrics for each bootstrap iteration. Values are reported as mean $\pm$ half-width of the 95\% confidence interval. $R^2$ measures the correlation between annotated and detected percentages across files for each class.

\begin{table}[ht!]
  \centering
  \scriptsize
  \caption{Comparison of species identification performance between DeepLabV3 and DINOv2 models across different taxonomical hierarchies on the BASE dataset test set (2019A2).}
  \label{tab:test_camera_species_experiments_comparison}
  \resizebox{0.95\textwidth}{!}{%
    \begin{tabular}{llccccc}
      \hline
      Experiment & Hierarchy  & Balanced Accuracy          & Precision (PPV)            & Recall (Sensitivity)       & F1 Score                   & R²                         \\
      \hline
      DeepLabV3  & Species    & $0.768 \pm 0.031$          & $0.559 \pm 0.064$          & $0.543 \pm 0.063$          & $0.534 \pm 0.057$          & $0.624 \pm 0.131$          \\
      DINOv2     & Species    & \textbf{$0.935 \pm 0.014$} & \textbf{$0.870 \pm 0.032$} & \textbf{$0.874 \pm 0.027$} & \textbf{$0.870 \pm 0.027$} & \textbf{$0.983 \pm 0.017$} \\
      DeepLabV3  & Genus      & $0.781 \pm 0.031$          & $0.594 \pm 0.067$          & $0.570 \pm 0.062$          & $0.567 \pm 0.058$          & $0.663 \pm 0.133$          \\
      DINOv2     & Genus      & \textbf{$0.942 \pm 0.013$} & \textbf{$0.891 \pm 0.029$} & \textbf{$0.887 \pm 0.025$} & \textbf{$0.888 \pm 0.025$} & \textbf{$0.987 \pm 0.014$} \\
      DeepLabV3  & Family     & $0.840 \pm 0.032$          & $0.712 \pm 0.060$          & $0.689 \pm 0.063$          & $0.691 \pm 0.055$          & $0.761 \pm 0.130$          \\
      DINOv2     & Family     & \textbf{$0.964 \pm 0.006$} & \textbf{$0.926 \pm 0.015$} & \textbf{$0.933 \pm 0.012$} & \textbf{$0.929 \pm 0.013$} & \textbf{$0.998 \pm 0.002$} \\
      DeepLabV3  & Order      & $0.868 \pm 0.020$          & $0.768 \pm 0.041$          & $0.747 \pm 0.039$          & $0.748 \pm 0.035$          & $0.817 \pm 0.100$          \\
      DINOv2     & Order      & \textbf{$0.969 \pm 0.004$} & \textbf{$0.947 \pm 0.007$} & \textbf{$0.945 \pm 0.008$} & \textbf{$0.946 \pm 0.007$} & \textbf{$0.999 \pm 0.001$} \\
      DeepLabV3  & Class      & $0.964 \pm 0.004$          & $0.951 \pm 0.006$          & $0.949 \pm 0.007$          & $0.950 \pm 0.006$          & $0.997 \pm 0.001$          \\
      DINOv2     & Class      & \textbf{$0.976 \pm 0.002$} & \textbf{$0.967 \pm 0.004$} & \textbf{$0.968 \pm 0.003$} & \textbf{$0.967 \pm 0.003$} & \textbf{$1.000 \pm 0.000$} \\
      DeepLabV3  & Vegetation & $0.970 \pm 0.002$          & $0.974 \pm 0.002$          & $0.970 \pm 0.003$          & $0.972 \pm 0.002$          & $0.998 \pm 0.000$          \\
      DINOv2     & Vegetation & \textbf{$0.977 \pm 0.001$} & \textbf{$0.977 \pm 0.002$} & \textbf{$0.977 \pm 0.002$} & \textbf{$0.977 \pm 0.002$} & \textbf{$1.000 \pm 0.000$} \\
      \hline
    \end{tabular}%
  }
\end{table}

Regarding damage assessment, Table~\ref{tab:test_camera_damage_experiments_comparison} shows the performance comparison between the proposed DINOv2-based model and the previous DeepLabV3-based model across damage classification tasks. The DINOv2 model demonstrates substantial improvements over the DeepLabV3 model in both damage class identification and healthy/damaged classification. For damage classes, the DINOv2 model achieves a balanced accuracy of $0.922 \pm 0.017$, a precision of $0.579 \pm 0.038$, a recall of $0.859 \pm 0.034$, an F1 score of $0.635 \pm 0.046$, and an $R^2$ of $0.844 \pm 0.074$. In contrast, the DeepLabV3 model records significantly lower metrics in all categories.

\begin{table}[ht!]
  \centering
  \scriptsize
  \caption{Comparison of damage assessment performance between DeepLabV3 and DINOv2 models on the BASE dataset test set (2019A2).}
  \label{tab:test_camera_damage_experiments_comparison}
  \resizebox{0.95\textwidth}{!}{%
    \begin{tabular}{llccccc}
      \hline
      Experiment & Hierarchy       & Balanced Accuracy          & Precision (PPV)            & Recall (Sensitivity)       & F1 Score                   & R²                         \\
      \hline
      DeepLabV3  & Damage Classes  & $0.811 \pm 0.032$          & $0.470 \pm 0.028$          & $0.652 \pm 0.065$          & $0.423 \pm 0.044$          & $0.538 \pm 0.087$          \\
      DINOv2     & Damage Classes  & \textbf{$0.922 \pm 0.017$} & \textbf{$0.579 \pm 0.038$} & \textbf{$0.859 \pm 0.034$} & \textbf{$0.635 \pm 0.046$} & \textbf{$0.844 \pm 0.074$} \\
      DeepLabV3  & Healthy/Damaged & $0.852 \pm 0.011$          & $0.705 \pm 0.015$          & $0.754 \pm 0.021$          & $0.592 \pm 0.034$          & $0.672 \pm 0.068$          \\
      DINOv2     & Healthy/Damaged & \textbf{$0.943 \pm 0.007$} & \textbf{$0.787 \pm 0.018$} & \textbf{$0.911 \pm 0.013$} & \textbf{$0.809 \pm 0.024$} & \textbf{$0.940 \pm 0.022$} \\
      \hline
    \end{tabular}%
  }
\end{table}

\subsubsection{Species identification}
Figures~\ref{fig:test_species_confusion_matrices} show the confusion matrices for species identification at different taxonomical hierarchies using both the DINOv2-based model and the DeepLabV3-based model on the BASE dataset test set (2019A2). The DINOv2 model exhibits higher accuracy and fewer misclassifications across all hierarchical levels compared to the DeepLabV3 model.

\begin{figure*}[ht!]
  \centering
  \footnotesize
  \includegraphics[width=14cm]{test_species_camera_confusion_matrices.jpg}
  \caption{Species identification confusion matrices for the DeepLabV3-based model (top) and the DINOv2-based model (bottom) on the BASE dataset test set (2019A2) evaluated at different taxonomical hierarchies: species, genus, family, order, class}
  \label{fig:test_species_confusion_matrices}
\end{figure*}

\begin{figure}[ht!]
  \centering
  \footnotesize
  \includegraphics[height=10cm]{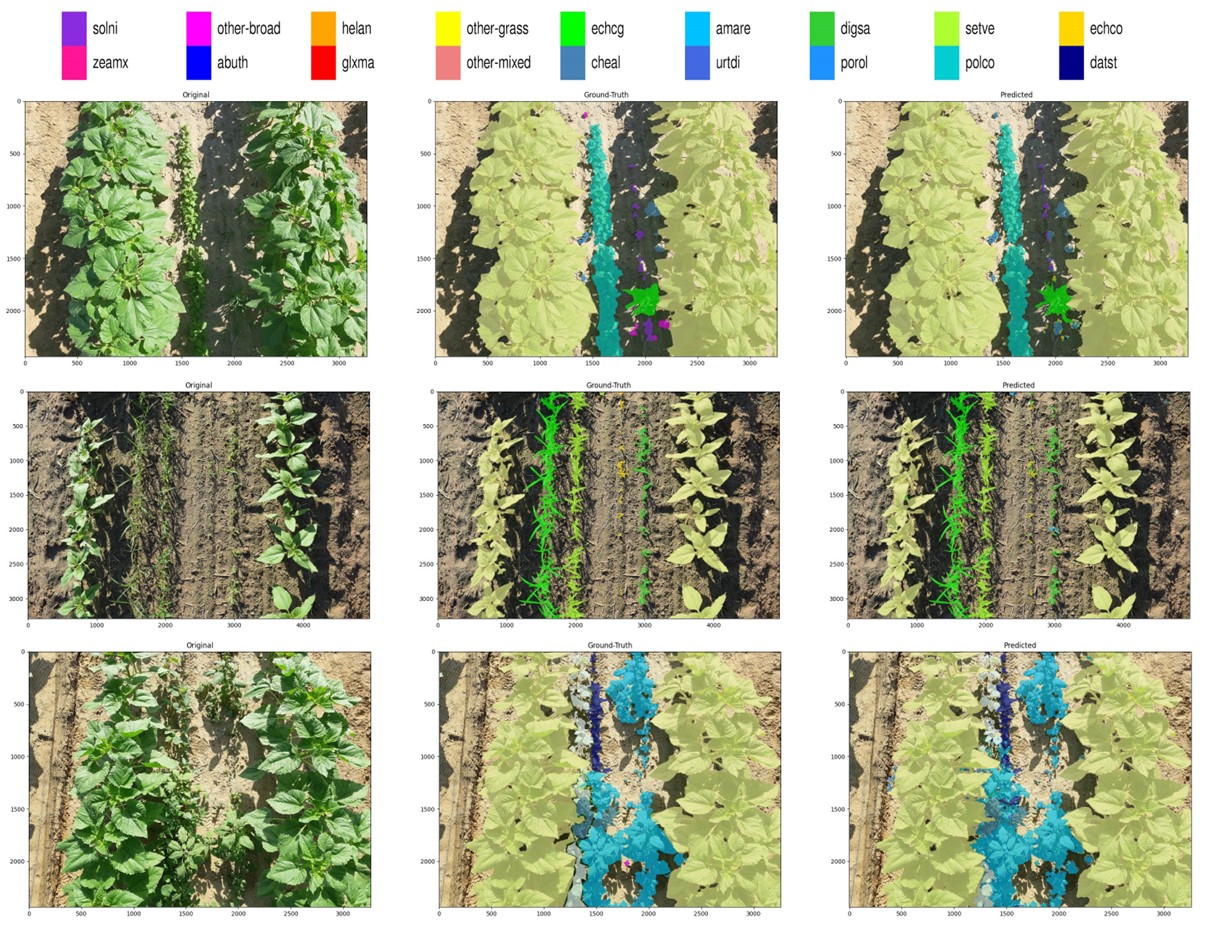}
  \caption{Algorithm species predictions for \textit{Helianthus annuus} field. Model trained and tested over the BASE dataset}
  \label{fig:species_results_test_helan}
\end{figure}

\begin{figure}[ht!]
  \centering
  \footnotesize
  \includegraphics[height=10cm]{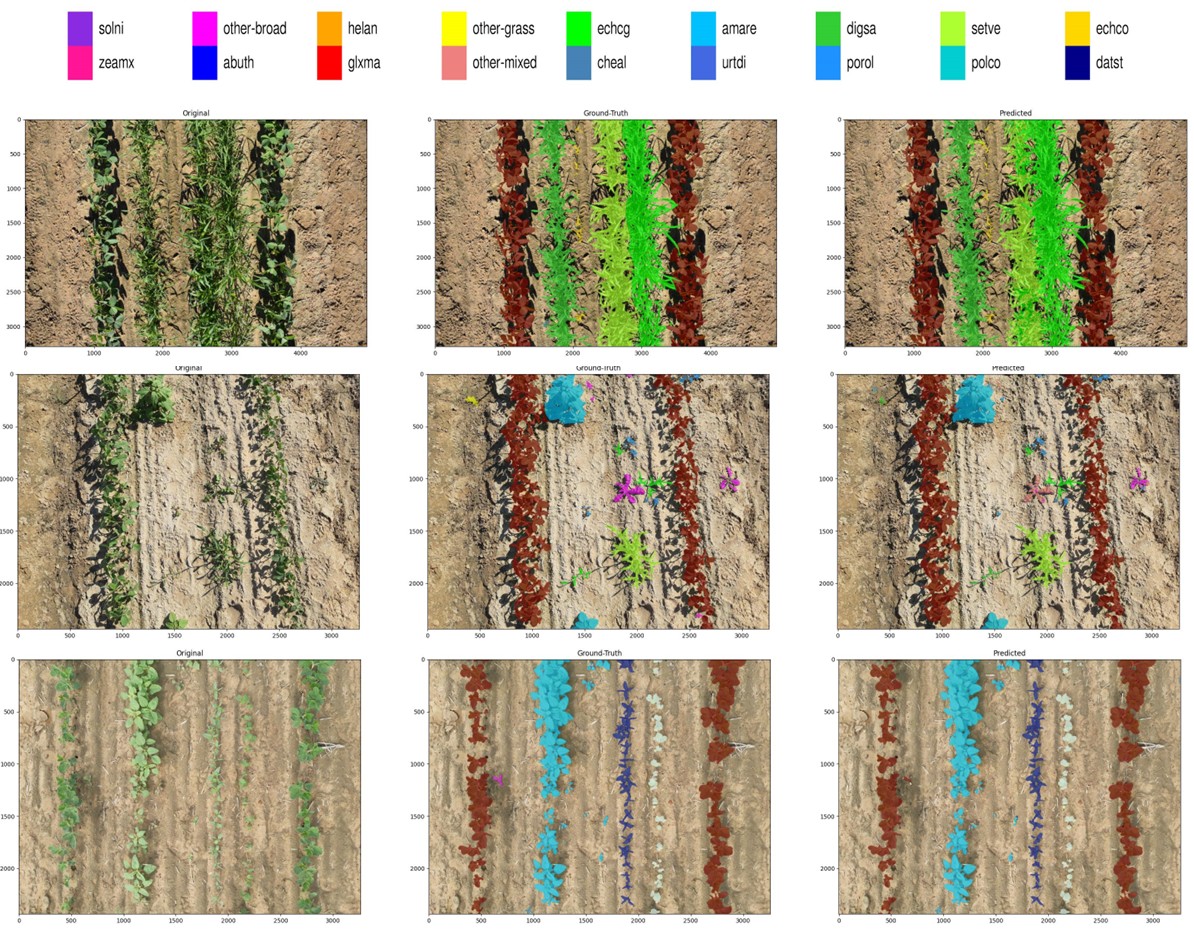}
  \caption{Algorithm species predictions for \textit{Glycine max} field. Model trained and tested over the BASE dataset}
  \label{fig:species_results_test_glxma}
\end{figure}

\begin{figure}[ht!]
  \centering
  \footnotesize
  \includegraphics[height=10cm]{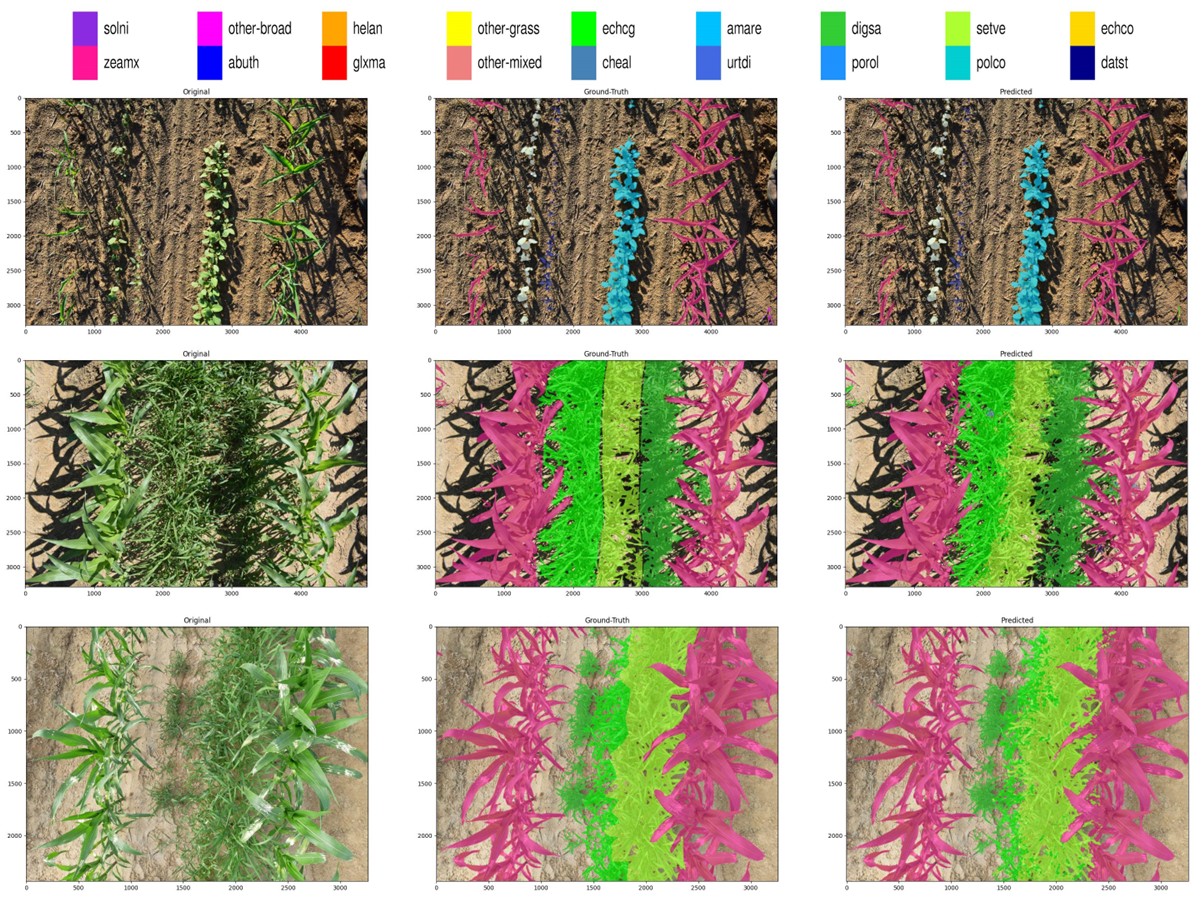}
  \caption{Algorithm species predictions for \textit{Zea mays} field. Model trained and tested over the BASE dataset}
  \label{fig:species_results_test_zeamx}
\end{figure}

Algorithm predictions are illustrated in Figures~\ref{fig:species_results_test_helan}, \ref{fig:species_results_test_glxma}, and \ref{fig:species_results_test_zeamx} for \textit{Helianthus annuus}, \textit{Glycine max}, and \textit{Zea mays} fields, respectively. These visualizations demonstrate the model's capability to accurately identify various species within different crop environments.

Table~\ref{tab:test_camera_species_experiments_perclass_dinov2_species} provides a detailed breakdown of per-class performance metrics for species identification using the DINOv2-based model on the BASE dataset test set (2019A2). The results indicate that the model performs exceptionally well across most species, with balanced accuracy values exceeding 0.9 for the majority of classes. However, certain species such as \textit{Echinochloa colona} and \textit{Solanum nigrum} exhibit relatively lower performance.

\begin{table}[ht!]
  \centering
  \scriptsize
  \caption{Per-class species identification performance metrics using the DINOv2-based model on the BASE dataset test set (2019A2).}
  \label{tab:test_camera_species_experiments_perclass_dinov2_species}
  \resizebox{0.95\textwidth}{!}{%
    \begin{tabular}{lccccc}
      \hline
      Class & Balanced Accuracy & Precision (PPV)   & Recall (Sensitivity) & F1 Score          & R²                \\
      \hline
      abuth & $0.951 \pm 0.007$ & $0.923 \pm 0.011$ & $0.901 \pm 0.014$    & $0.912 \pm 0.012$ & $0.997 \pm 0.002$ \\
      amare & $0.969 \pm 0.006$ & $0.940 \pm 0.014$ & $0.939 \pm 0.011$    & $0.939 \pm 0.011$ & $0.998 \pm 0.003$ \\
      cheal & $0.936 \pm 0.015$ & $0.842 \pm 0.045$ & $0.873 \pm 0.030$    & $0.857 \pm 0.034$ & $0.993 \pm 0.011$ \\
      datst & $0.960 \pm 0.007$ & $0.908 \pm 0.014$ & $0.920 \pm 0.014$    & $0.914 \pm 0.013$ & $0.999 \pm 0.001$ \\
      digsa & $0.903 \pm 0.044$ & $0.837 \pm 0.046$ & $0.808 \pm 0.087$    & $0.822 \pm 0.067$ & $0.973 \pm 0.030$ \\
      echcg & $0.944 \pm 0.008$ & $0.846 \pm 0.018$ & $0.891 \pm 0.015$    & $0.868 \pm 0.013$ & $0.988 \pm 0.006$ \\
      echco & $0.844 \pm 0.020$ & $0.572 \pm 0.074$ & $0.688 \pm 0.040$    & $0.624 \pm 0.052$ & $0.920 \pm 0.063$ \\
      glxma & $0.978 \pm 0.002$ & $0.957 \pm 0.004$ & $0.958 \pm 0.003$    & $0.958 \pm 0.003$ & $0.999 \pm 0.001$ \\
      helan & $0.988 \pm 0.002$ & $0.984 \pm 0.003$ & $0.977 \pm 0.004$    & $0.980 \pm 0.003$ & $1.000 \pm 0.000$ \\
      misc  & $0.977 \pm 0.001$ & $0.992 \pm 0.001$ & $0.992 \pm 0.001$    & $0.992 \pm 0.001$ & $1.000 \pm 0.000$ \\
      polco & $0.966 \pm 0.015$ & $0.952 \pm 0.025$ & $0.933 \pm 0.030$    & $0.942 \pm 0.027$ & $1.000 \pm 0.001$ \\
      porol & $0.926 \pm 0.011$ & $0.753 \pm 0.056$ & $0.851 \pm 0.022$    & $0.799 \pm 0.039$ & $0.992 \pm 0.012$ \\
      setve & $0.910 \pm 0.016$ & $0.847 \pm 0.043$ & $0.822 \pm 0.032$    & $0.834 \pm 0.034$ & $0.983 \pm 0.009$ \\
      solni & $0.799 \pm 0.045$ & $0.736 \pm 0.117$ & $0.597 \pm 0.091$    & $0.659 \pm 0.095$ & $0.902 \pm 0.122$ \\
      zeamx & $0.977 \pm 0.004$ & $0.959 \pm 0.008$ & $0.955 \pm 0.008$    & $0.957 \pm 0.007$ & $1.000 \pm 0.000$ \\
      \hline
    \end{tabular}%
  }
\end{table}



In summary, the proposed DINOv2-based model demonstrates superior performance in species identification compared to the previous DeepLabV3-based model across various taxonomical hierarchies. The detailed per-class analysis further highlights the model's effectiveness in accurately identifying a wide range of species within the agricultural context under no domain shift conditions.

\subsubsection{Damage identification}

\begin{figure*}[ht!]
  \centering
  \footnotesize
  \includegraphics[width=10cm]{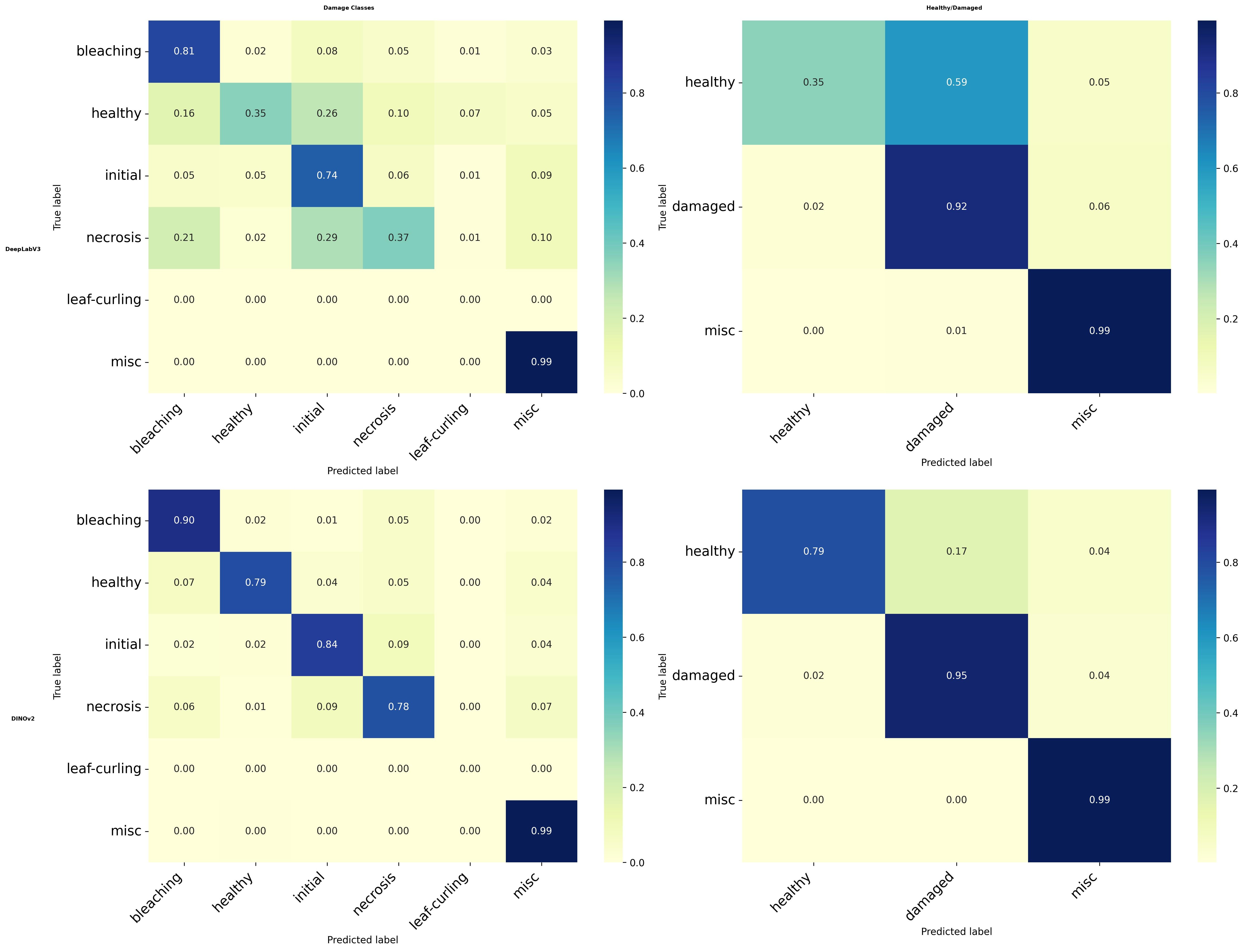}
  \caption{Damage identification confusion matrices for the DINOv2-based model and the DeepLabV3-based model on the BASE dataset test set (2019A2) evaluated at different damage hierarchies: damage classes (bleaching, initial, necrosis) and healthy/damaged}
  \label{fig:test_damage_confusion_matrices}
\end{figure*}

Figures~\ref{fig:test_damage_confusion_matrices} show the confusion matrices for damage identification using both the DINOv2-based model (bottom) and the DeepLabV3-based model (top) on the BASE dataset test set (2019A2). The DINOv2 model exhibits higher accuracy and fewer misclassifications across all damage hierarchies compared to the DeepLabV3 model.

\begin{figure}[ht!]
  \centering
  \footnotesize
  \includegraphics[height=10cm]{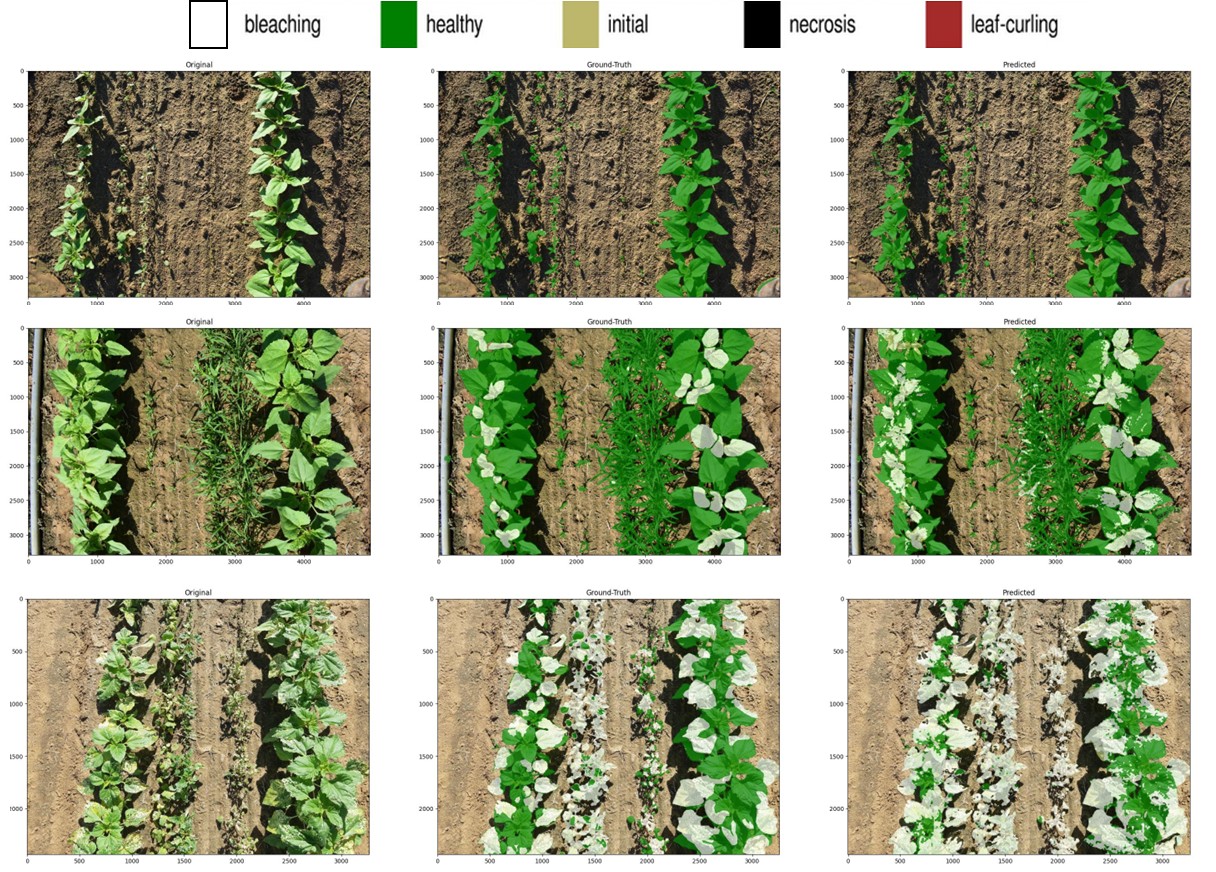}
  \caption{Algorithm damage predictions for \textit{Helianthus annuus} field. Model trained and tested over the BASE dataset}
  \label{fig:damage_results_test_helan}
\end{figure}

\begin{figure}[ht!]
  \centering
  \footnotesize
  \includegraphics[height=10cm]{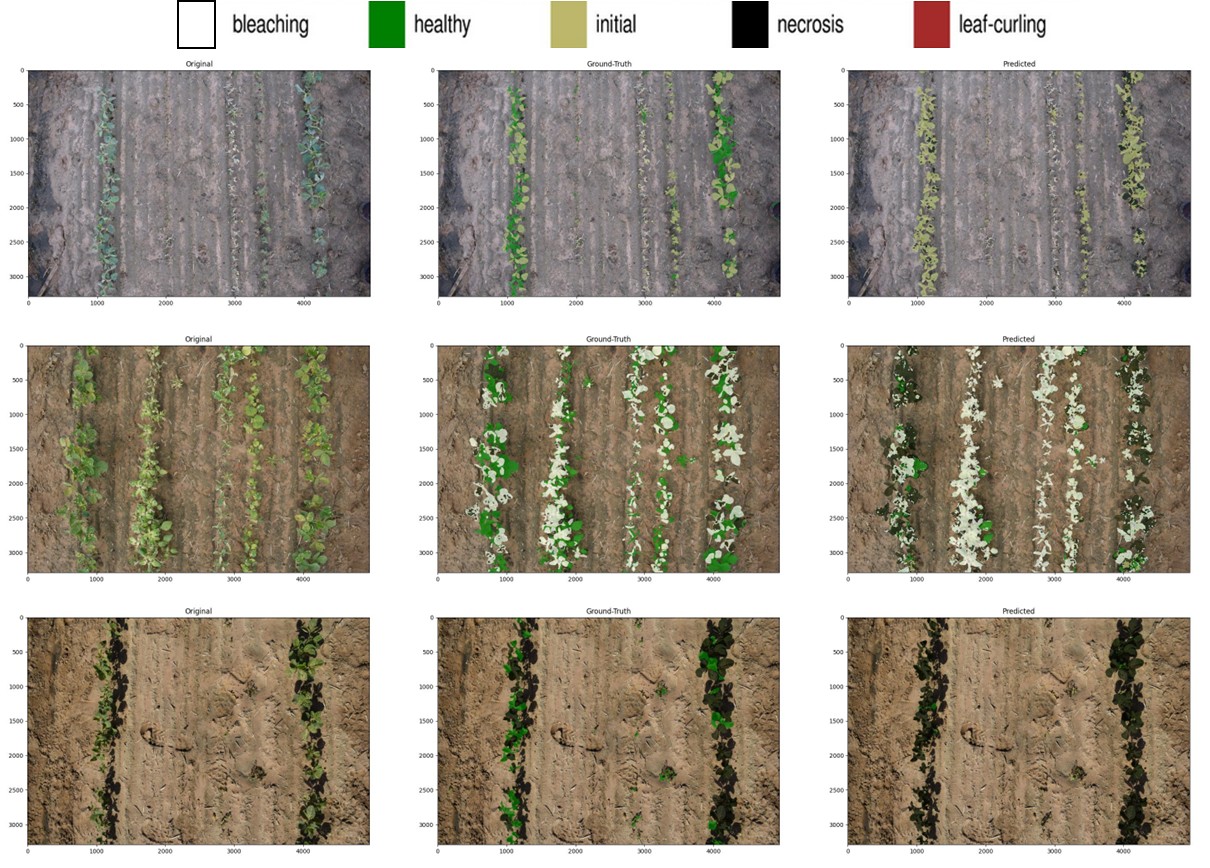}
  \caption{Algorithm damage predictions for \textit{Glycine max} field. Model trained and tested over the BASE dataset}
  \label{fig:damage_results_test_glxma}
\end{figure}

\begin{figure}[ht!]
  \centering
  \footnotesize
  \includegraphics[height=10cm]{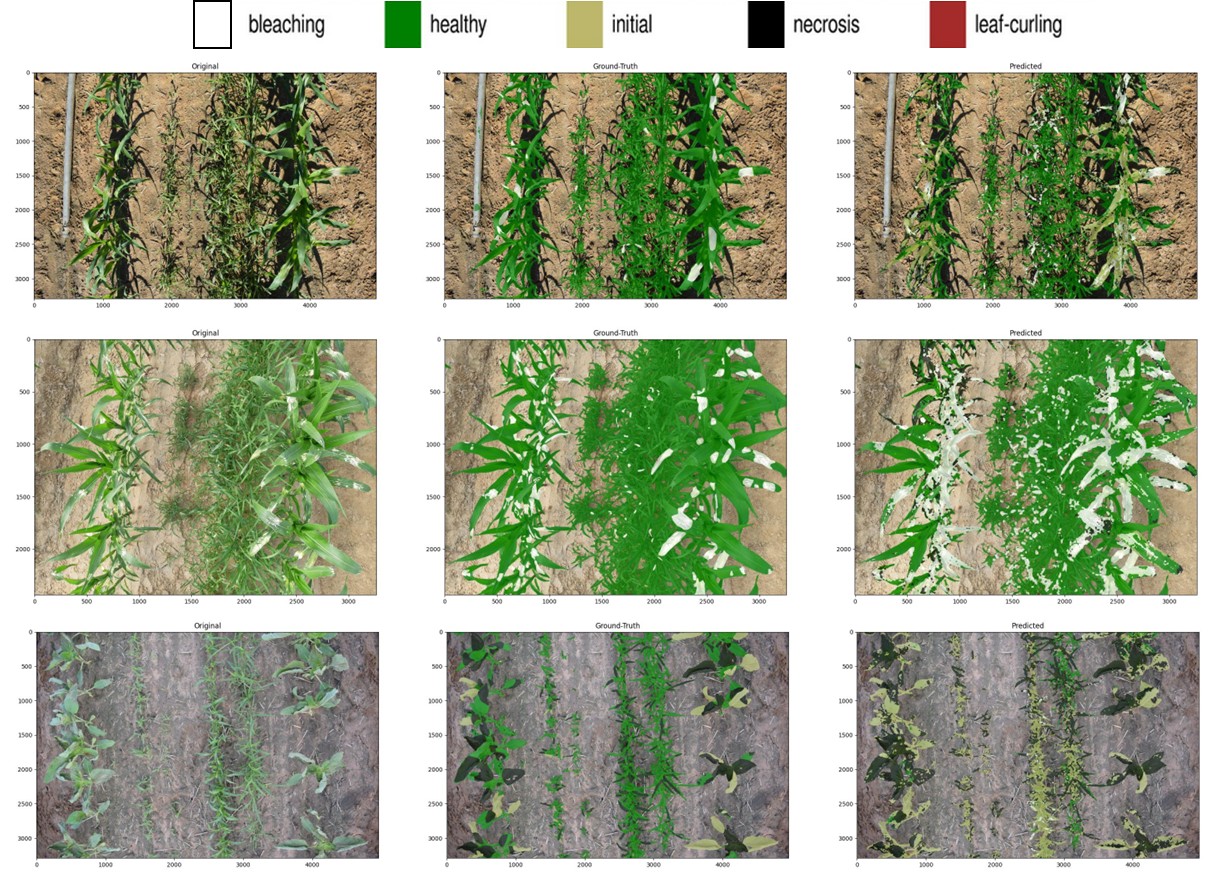}
  \caption{Algorithm damage predictions for \textit{Zea mays} field. Model trained and tested over the BASE dataset}
  \label{fig:damage_results_test_zeamx}
\end{figure}

Algorithm predictions are illustrated in Figures~\ref{fig:damage_results_test_helan}, \ref{fig:damage_results_test_glxma}, and \ref{fig:damage_results_test_zeamx} for \textit{Helianthus annuus}, \textit{Glycine max}, and \textit{Zea mays} fields, respectively. These visualizations demonstrate the model's capability to accurately assess various damage types within different crop environments.

Table~\ref{tab:test_camera_damage_experiments_perclass_dinov2_damage_classes} provides a detailed breakdown of per-class performance metrics for damage identification using the DINOv2-based model on the BASE dataset test set (2019A2). The results indicate that the model performs well across most damage classes, with balanced accuracy values exceeding 0.9 for bleaching and misc classes. However, certain damage types such as initial and necrosis exhibit relatively lower performance. By definition, initial damage may be more challenging to detect due to its subtle visual cues compared to more pronounced damage types like bleaching and necrosis.

\begin{table}[ht!]
  \centering
  \scriptsize
  \caption{Per-class performance metrics for damage identification using the DINOv2-based model on the BASE dataset test set (2019A2) across different damage classes.}
  \label{tab:test_camera_damage_experiments_perclass_dinov2_damage_classes}
  \resizebox{0.95\textwidth}{!}{%
    \begin{tabular}{lccccc}
      \hline
      Class     & Balanced Accuracy & Precision (PPV)   & Recall (Sensitivity) & F1 Score          & R²                \\
      \hline
      bleaching & $0.942 \pm 0.014$ & $0.453 \pm 0.066$ & $0.897 \pm 0.028$    & $0.601 \pm 0.064$ & $0.927 \pm 0.038$ \\
      healthy   & $0.893 \pm 0.015$ & $0.968 \pm 0.003$ & $0.791 \pm 0.030$    & $0.871 \pm 0.019$ & $0.946 \pm 0.021$ \\
      initial   & $0.914 \pm 0.027$ & $0.138 \pm 0.057$ & $0.835 \pm 0.054$    & $0.236 \pm 0.085$ & $0.609 \pm 0.197$ \\
      misc      & $0.976 \pm 0.001$ & $0.991 \pm 0.001$ & $0.993 \pm 0.001$    & $0.992 \pm 0.001$ & $1.000 \pm 0.000$ \\
      necrosis  & $0.885 \pm 0.029$ & $0.343 \pm 0.060$ & $0.780 \pm 0.058$    & $0.476 \pm 0.062$ & $0.740 \pm 0.116$ \\
      \hline
    \end{tabular}%
  }
\end{table}



In summary, the proposed DINOv2-based model demonstrates superior performance in damage identification compared to the previous DeepLabV3-based model. The detailed per-class analysis further highlights the model's effectiveness in accurately identifying various damage types within the agricultural context under no domain shift conditions.

\subsection{Evaluation on the REALITY Dataset: Assessing Domain Shift Robustness}
\label{ssec:results_reality}

In the REALITY dataset, factors such as geographical region, date, climate, drought, device, and weather, among others, can produce data biases in the organisms. To evaluate the real performance of the algorithm, in this experiment the algorithm was trained with the BASE dataset (generated between 2018--2020 in Germany and Spain) and tested with the REALITY dataset (generated in 2023 in Germany, Spain, and the U.S.). This ensures unbiased testing that evaluates the algorithm's performance under unseen scenarios using the same type of sensors (mobile phones and digital cameras) as used in the BASE dataset.

Table~\ref{tab:reality_check_camera_species_experiments_comparison} presents a comprehensive comparison of the proposed DINOv2-based model against the previous DeepLabV3-based model across different taxonomical hierarchies for species identification on the REALITY dataset. The DINOv2 model consistently outperforms the DeepLabV3 model at all hierarchical levels, from species to vegetation. Notably, at the species level, the DINOv2 model achieves a balanced accuracy of $0.898 \pm 0.050$, a precision of $0.741 \pm 0.117$, a recall of $0.803 \pm 0.100$, an F1 score of $0.766 \pm 0.102$, and an $R^2$ of $0.883 \pm 0.135$. In contrast, the DeepLabV3 model records significantly lower metrics in all categories.

\begin{table}[ht!]
  \centering
  \scriptsize
  \caption{Comparison of species identification performance between DeepLabV3 and DINOv2 models across different taxonomical hierarchies on the REALITY dataset.}
  \label{tab:reality_check_camera_species_experiments_comparison}
  \resizebox{0.95\textwidth}{!}{%
    \begin{tabular}{llccccc}
      \hline
      Experiment & Hierarchy  & Balanced Accuracy          & Precision (PPV)            & Recall (Sensitivity)       & F1 Score                   & R²                         \\
      \hline
      DeepLabV3  & Species    & $0.645 \pm 0.054$          & $0.375 \pm 0.111$          & $0.306 \pm 0.108$          & $0.243 \pm 0.085$          & $0.193 \pm 0.159$          \\
      DINOv2     & Species    & \textbf{$0.898 \pm 0.050$} & \textbf{$0.741 \pm 0.117$} & \textbf{$0.803 \pm 0.100$} & \textbf{$0.766 \pm 0.102$} & \textbf{$0.883 \pm 0.135$} \\
      DeepLabV3  & Genus      & $0.648 \pm 0.054$          & $0.366 \pm 0.105$          & $0.311 \pm 0.108$          & $0.248 \pm 0.085$          & $0.191 \pm 0.157$          \\
      DINOv2     & Genus      & \textbf{$0.899 \pm 0.049$} & \textbf{$0.740 \pm 0.117$} & \textbf{$0.804 \pm 0.097$} & \textbf{$0.765 \pm 0.100$} & \textbf{$0.883 \pm 0.139$} \\
      DeepLabV3  & Family     & $0.688 \pm 0.050$          & $0.430 \pm 0.068$          & $0.400 \pm 0.100$          & $0.319 \pm 0.059$          & $0.317 \pm 0.136$          \\
      DINOv2     & Family     & \textbf{$0.907 \pm 0.039$} & \textbf{$0.778 \pm 0.096$} & \textbf{$0.823 \pm 0.077$} & \textbf{$0.796 \pm 0.081$} & \textbf{$0.942 \pm 0.102$} \\
      DeepLabV3  & Order      & $0.716 \pm 0.046$          & $0.525 \pm 0.061$          & $0.461 \pm 0.092$          & $0.414 \pm 0.067$          & $0.403 \pm 0.143$          \\
      DINOv2     & Order      & \textbf{$0.939 \pm 0.020$} & \textbf{$0.851 \pm 0.050$} & \textbf{$0.890 \pm 0.041$} & \textbf{$0.868 \pm 0.040$} & \textbf{$0.961 \pm 0.029$} \\
      DeepLabV3  & Class      & $0.880 \pm 0.023$          & $0.839 \pm 0.041$          & $0.800 \pm 0.044$          & $0.808 \pm 0.041$          & $0.754 \pm 0.124$          \\
      DINOv2     & Class      & \textbf{$0.955 \pm 0.008$} & \textbf{$0.911 \pm 0.017$} & \textbf{$0.936 \pm 0.013$} & \textbf{$0.923 \pm 0.013$} & \textbf{$0.968 \pm 0.021$} \\
      DeepLabV3  & Vegetation & $0.957 \pm 0.007$          & $0.942 \pm 0.014$          & $0.957 \pm 0.010$          & $0.949 \pm 0.009$          & $0.917 \pm 0.054$          \\
      DINOv2     & Vegetation & \textbf{$0.960 \pm 0.005$} & \textbf{$0.942 \pm 0.011$} & \textbf{$0.960 \pm 0.008$} & \textbf{$0.951 \pm 0.008$} & \textbf{$0.963 \pm 0.023$} \\
      \hline
    \end{tabular}%
  }
\end{table}

For damage assessment, Table~\ref{tab:reality_check_camera_damage_experiments_comparison} shows the performance comparison between the proposed DINOv2-based model and the previous DeepLabV3-based model across damage classification tasks on the REALITY dataset. The DINOv2 model demonstrates improvements over the DeepLabV3 model in both damage class identification and healthy/damaged classification. For damage classes, the DINOv2 model achieves a balanced accuracy of $0.732 \pm 0.052$, a precision of $0.436 \pm 0.090$, a recall of $0.492 \pm 0.103$, an F1 score of $0.402 \pm 0.061$, and an $R^2$ of $0.522 \pm 0.207$. In contrast, the DeepLabV3 model records lower metrics in all categories.

\begin{table}[ht!]
  \centering
  \scriptsize
  \caption{Comparison of damage identification performance between DeepLabV3 and DINOv2 models across different damage hierarchies on the REALITY dataset.}
  \label{tab:reality_check_camera_damage_experiments_comparison}
  \resizebox{0.95\textwidth}{!}{%
    \begin{tabular}{llccccc}
      \hline
      Experiment & Hierarchy       & Balanced Accuracy          & Precision (PPV)            & Recall (Sensitivity)       & F1 Score                   & R²                         \\
      \hline
      DeepLabV3  & Damage Classes  & $0.713 \pm 0.041$          & $0.379 \pm 0.045$          & $0.465 \pm 0.080$          & $0.347 \pm 0.054$          & $0.418 \pm 0.198$          \\
      DINOv2     & Damage Classes  & \textbf{$0.732 \pm 0.052$} & \textbf{$0.436 \pm 0.090$} & \textbf{$0.492 \pm 0.103$} & \textbf{$0.402 \pm 0.061$} & \textbf{$0.522 \pm 0.207$} \\
      DeepLabV3  & Healthy/Damaged & $0.832 \pm 0.035$          & $0.702 \pm 0.045$          & \textbf{$0.738 \pm 0.064$} & $0.660 \pm 0.066$          & \textbf{$0.690 \pm 0.150$} \\
      DINOv2     & Healthy/Damaged & \textbf{$0.841 \pm 0.032$} & \textbf{$0.724 \pm 0.045$} & $0.736 \pm 0.060$          & \textbf{$0.726 \pm 0.036$} & $0.675 \pm 0.138$          \\
      \hline
    \end{tabular}%
  }
\end{table}

It is evident that the performance of both models degrades when evaluated under domain shift conditions. However, the DINOv2-based model shows a smaller performance drop compared to the DeepLabV3-based model, indicating better robustness to domain shifts. The drop in F1-score at species level is 0.266 for the DeepLabV3-based model and 0.102 for the DINOv2-based model. For damage assessment, the F1-score drop for damage classes is 0.233 for both models. However, for the healthy/damaged classification, the DINOv2-based model shows a smaller drop (0.083) compared to the DeepLabV3-based model (0.232). These results highlight the advantages of using self-supervised foundation models like DINOv2 for agricultural applications, particularly in scenarios where domain shifts are prevalent.

\subsubsection{Species identification}
\label{sssec:results_reality_species}

Figures~\ref{fig:reality_species_camera_confusion_matrices} show the confusion matrices for species identification at different taxonomical hierarchies using both the DINOv2-based model and the DeepLabV3-based model on the REALITY dataset. The DINOv2 model exhibits higher accuracy and fewer misclassifications across all hierarchical levels compared to the DeepLabV3 model.

\begin{figure*}[ht!]
  \centering
  \footnotesize
  \includegraphics[width=14cm]{reality_check_species_camera_confusion_matrices.jpg}
  \caption{Species identification confusion matrices for the DINOv2-based model (bottom) and the DeepLabV3-based model (top) on the REALITY dataset evaluated at different taxonomical hierarchies: species, genus, family, order, class}
  \label{fig:reality_species_camera_confusion_matrices}
\end{figure*}

\begin{figure}[ht!]
  \centering
  \footnotesize
  \includegraphics[height=10cm]{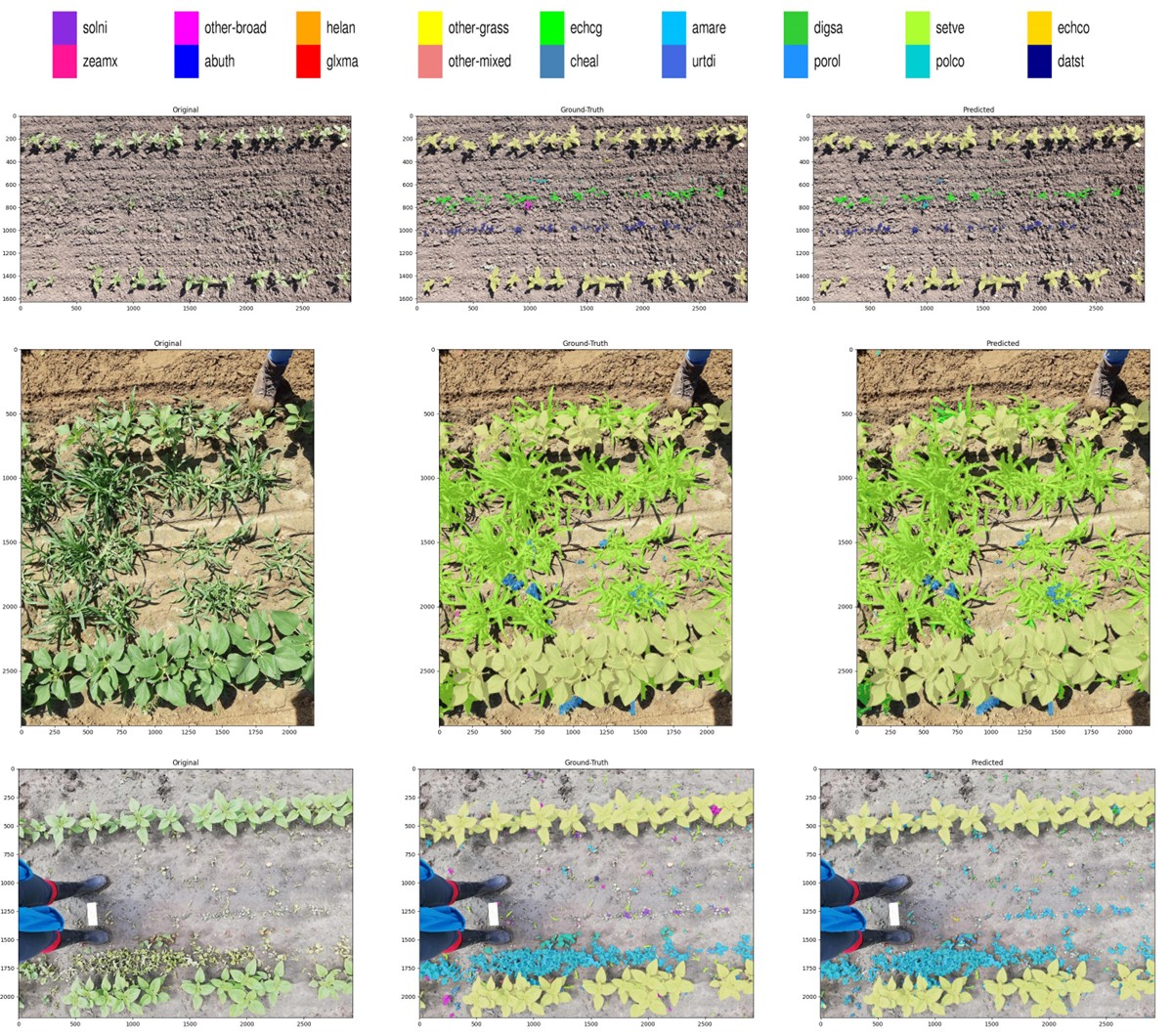}
  \caption{Algorithm species predictions for \textit{Helianthus annuus} field. Model trained on the BASE dataset and tested on the REALITY dataset}
  \label{fig:species_results_reality_helan}
\end{figure}

\begin{figure}[ht!]
  \centering
  \footnotesize
  \includegraphics[height=10cm]{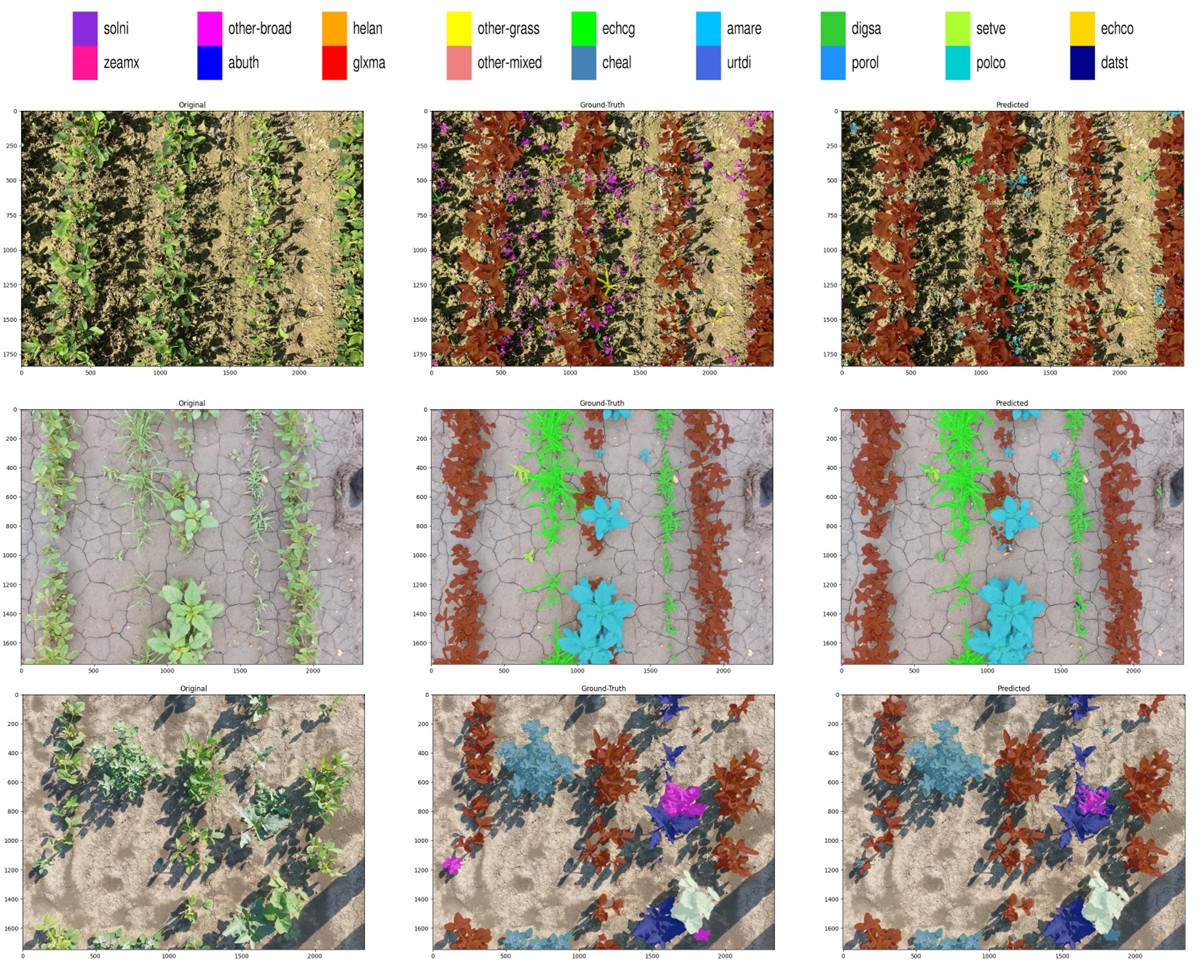}
  \caption{Algorithm species predictions for \textit{Glycine max} field. Model trained on the BASE dataset and tested on the REALITY dataset}
  \label{fig:species_results_reality_glxma}
\end{figure}

\begin{figure}[ht!]
  \centering
  \footnotesize
  \includegraphics[height=10cm]{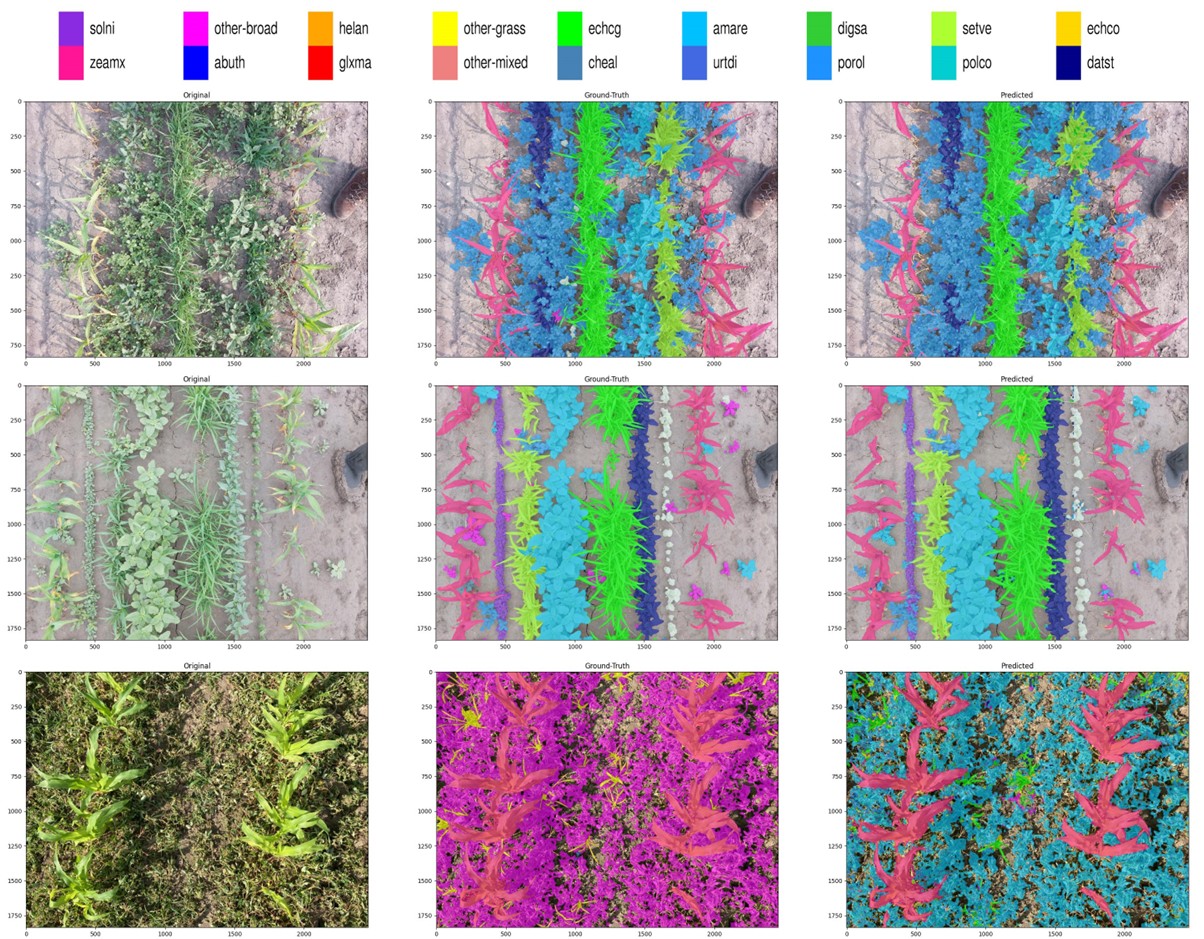}
  \caption{Algorithm species predictions for \textit{Zea mays} field. Model trained on the BASE dataset and tested on the REALITY dataset}
  \label{fig:species_results_reality_zeamx}
\end{figure}

Algorithm predictions are illustrated in Figures~\ref{fig:species_results_reality_helan}, \ref{fig:species_results_reality_glxma}, and \ref{fig:species_results_reality_zeamx} for \textit{Helianthus annuus}, \textit{Glycine max}, and \textit{Zea mays} fields, respectively. These visualizations demonstrate the model's capability to accurately identify various species within different crop environments.

Table~\ref{tab:reality_check_camera_species_experiments_perclass_dinov2_species} provides a detailed breakdown of per-class performance metrics for species identification using the DINOv2-based model on the REALITY dataset. The results indicate that the model performs well across most species, with balanced accuracy values exceeding 0.9 for several classes. However, certain species such as \textit{Digitaria sanguinalis} and \textit{Polypogon convolutus} exhibit relatively lower performance.

\begin{table}[ht!]
  \centering
  \scriptsize
  \caption{Per-class species identification performance metrics using the DINOv2-based model on the REALITY dataset.}
  \label{tab:reality_check_camera_species_experiments_perclass_dinov2_species}
  \resizebox{0.95\textwidth}{!}{%
    \begin{tabular}{lccccc}
      \hline
      Class & Balanced Accuracy & Precision (PPV)   & Recall (Sensitivity) & F1 Score          & R²                \\
      \hline
      abuth & $0.876 \pm 0.062$ & $0.776 \pm 0.130$ & $0.752 \pm 0.125$    & $0.761 \pm 0.102$ & $0.923 \pm 0.044$ \\
      amare & $0.925 \pm 0.028$ & $0.657 \pm 0.122$ & $0.857 \pm 0.057$    & $0.742 \pm 0.091$ & $0.770 \pm 0.183$ \\
      cheal & $0.952 \pm 0.019$ & $0.693 \pm 0.124$ & $0.909 \pm 0.040$    & $0.785 \pm 0.093$ & $0.984 \pm 0.024$ \\
      datst & $0.935 \pm 0.035$ & $0.723 \pm 0.119$ & $0.870 \pm 0.069$    & $0.789 \pm 0.087$ & $0.976 \pm 0.036$ \\
      digsa & $0.797 \pm 0.158$ & $0.411 \pm 0.198$ & $0.596 \pm 0.315$    & $0.474 \pm 0.222$ & $0.267 \pm 0.435$ \\
      echcg & $0.862 \pm 0.045$ & $0.706 \pm 0.092$ & $0.728 \pm 0.091$    & $0.716 \pm 0.082$ & $0.848 \pm 0.119$ \\
      glxma & $0.963 \pm 0.009$ & $0.924 \pm 0.011$ & $0.931 \pm 0.018$    & $0.928 \pm 0.012$ & $0.997 \pm 0.003$ \\
      helan & $0.976 \pm 0.008$ & $0.963 \pm 0.006$ & $0.955 \pm 0.016$    & $0.959 \pm 0.009$ & $0.999 \pm 0.001$ \\
      misc  & $0.960 \pm 0.005$ & $0.986 \pm 0.003$ & $0.971 \pm 0.006$    & $0.978 \pm 0.004$ & $0.963 \pm 0.024$ \\
      polco & $0.710 \pm 0.131$ & $0.480 \pm 0.351$ & $0.421 \pm 0.263$    & $0.442 \pm 0.293$ & $0.796 \pm 0.478$ \\
      porol & $0.866 \pm 0.071$ & $0.587 \pm 0.134$ & $0.736 \pm 0.143$    & $0.652 \pm 0.124$ & $0.933 \pm 0.203$ \\
      setve & $0.884 \pm 0.064$ & $0.732 \pm 0.210$ & $0.772 \pm 0.129$    & $0.749 \pm 0.172$ & $0.931 \pm 0.283$ \\
      solni & $0.905 \pm 0.050$ & $0.825 \pm 0.111$ & $0.811 \pm 0.101$    & $0.817 \pm 0.091$ & $0.978 \pm 0.039$ \\
      zeamx & $0.959 \pm 0.013$ & $0.918 \pm 0.015$ & $0.920 \pm 0.027$    & $0.919 \pm 0.018$ & $0.995 \pm 0.004$ \\
      \hline
    \end{tabular}%
  }
\end{table}

Classification confusion matrices at different taxonomical levels can be seen in Figure~\ref{fig:reality_species_camera_confusion_matrices}. The same clusters observed in the species confusion matrices are present. At the family and order levels, minor confusion clusters are detected. The Malvaceae family is mixed with the Amaranthaceae and Polygonaceae families. When analyzed at the order level, accuracy is greater than 0.85 except for the Malvales, which are mixed with the Caryophyllales.

The classification accuracy for broadleaf species and grass-leaf species is above 0.89 in all cases, and we found no relevant misclassifications among them. They are misclassified as miscellaneous (misc). However, this is due to the elevated percentage of damaged vegetation included in the REALITY dataset, as the misc category accounts for human annotation errors and plants with a high percentage of damage.

The results in Table~\ref{tab:reality_check_camera_species_experiments_perclass_dinov2_species} demonstrate the model's ability to generalize across different locations and species. Notably, the model exhibits high accuracy for most species, with only a few species exhibiting lower performance. The species exhibiting lower performance are \textit{Echinochloa crus-galli}, \textit{Digitaria sanguinalis}, and \textit{Polypogon convolutus}. Additionally, there are small misclassification clusters, including one involving grass species such as \textit{Echinochloa crus-galli} and \textit{Digitaria sanguinalis}, and another involving broad-leaved species such as \textit{Polygonum convolvulus}, \textit{Amaranthus retroflexus}, \textit{Abutilon theophrasti}, \textit{Galinsoga parviflora}, and \textit{Chenopodium album}. Despite these issues, the model correctly identified the majority of species.



When analyzing the performance per location, Table~\ref{tab:reality_check_camera_species_experiments_perlocation_dinov2_species} presents the results for Germany (DE), Spain (ES), and the United States (US). The DINOv2-based model consistently outperforms the DeepLabV3-based model across all locations. Notably, the model achieves the highest balanced accuracy in Spain ($0.913 \pm 0.047$) and the lowest in the United States ($0.731 \pm 0.080$). This is expected, as the training BASE dataset was acquired in Germany and Spain, with no data from the U.S. being included. The results highlight the model's robustness and adaptability to different geographical regions, although performance may vary based on the similarity of the test data to the training data.

\begin{table}[ht!]
  \centering
  \scriptsize
  \caption{Per-location species identification performance metrics using the DINOv2-based model on the REALITY dataset.}
  \label{tab:reality_check_camera_species_experiments_perlocation_dinov2_species}
  \resizebox{0.95\textwidth}{!}{%
    \begin{tabular}{llccccc}
      \hline
      Experiment & Location & Balanced Accuracy & Precision (PPV)   & Recall (Sensitivity) & F1 Score          & R²                \\
      \hline
      DINOv2     & DE       & $0.831 \pm 0.059$ & $0.587 \pm 0.142$ & $0.668 \pm 0.118$    & $0.609 \pm 0.134$ & $0.906 \pm 0.251$ \\
      DINOv2     & ES       & $0.913 \pm 0.047$ & $0.797 \pm 0.107$ & $0.831 \pm 0.094$    & $0.809 \pm 0.100$ & $0.919 \pm 0.085$ \\
      DINOv2     & US       & $0.731 \pm 0.080$ & $0.436 \pm 0.163$ & $0.471 \pm 0.160$    & $0.434 \pm 0.143$ & $0.566 \pm 0.201$ \\
      DeepLabV3  & DE       & $0.624 \pm 0.058$ & $0.346 \pm 0.094$ & $0.266 \pm 0.115$    & $0.212 \pm 0.081$ & $0.262 \pm 0.210$ \\
      DeepLabV3  & ES       & $0.649 \pm 0.071$ & $0.420 \pm 0.170$ & $0.316 \pm 0.141$    & $0.288 \pm 0.125$ & $0.355 \pm 0.269$ \\
      DeepLabV3  & US       & $0.575 \pm 0.034$ & $0.307 \pm 0.064$ & $0.163 \pm 0.063$    & $0.169 \pm 0.048$ & $0.211 \pm 0.156$ \\
      \hline
      DINOv2     & All      & $0.898 \pm 0.050$ & $0.741 \pm 0.116$ & $0.802 \pm 0.100$    & $0.765 \pm 0.100$ & $0.883 \pm 0.134$ \\
      DeepLabV3  & All      & $0.644 \pm 0.054$ & $0.374 \pm 0.111$ & $0.305 \pm 0.108$    & $0.242 \pm 0.086$ & $0.193 \pm 0.159$ \\
      \hline
    \end{tabular}%
  }
\end{table}

In summary, the DINOv2-based model demonstrates superior performance in species identification compared to the previous DeepLabV3-based model on the REALITY dataset.

\subsubsection{Damage identification}

\begin{figure*}[ht!]
  \centering
  \footnotesize
  \includegraphics[width=10cm]{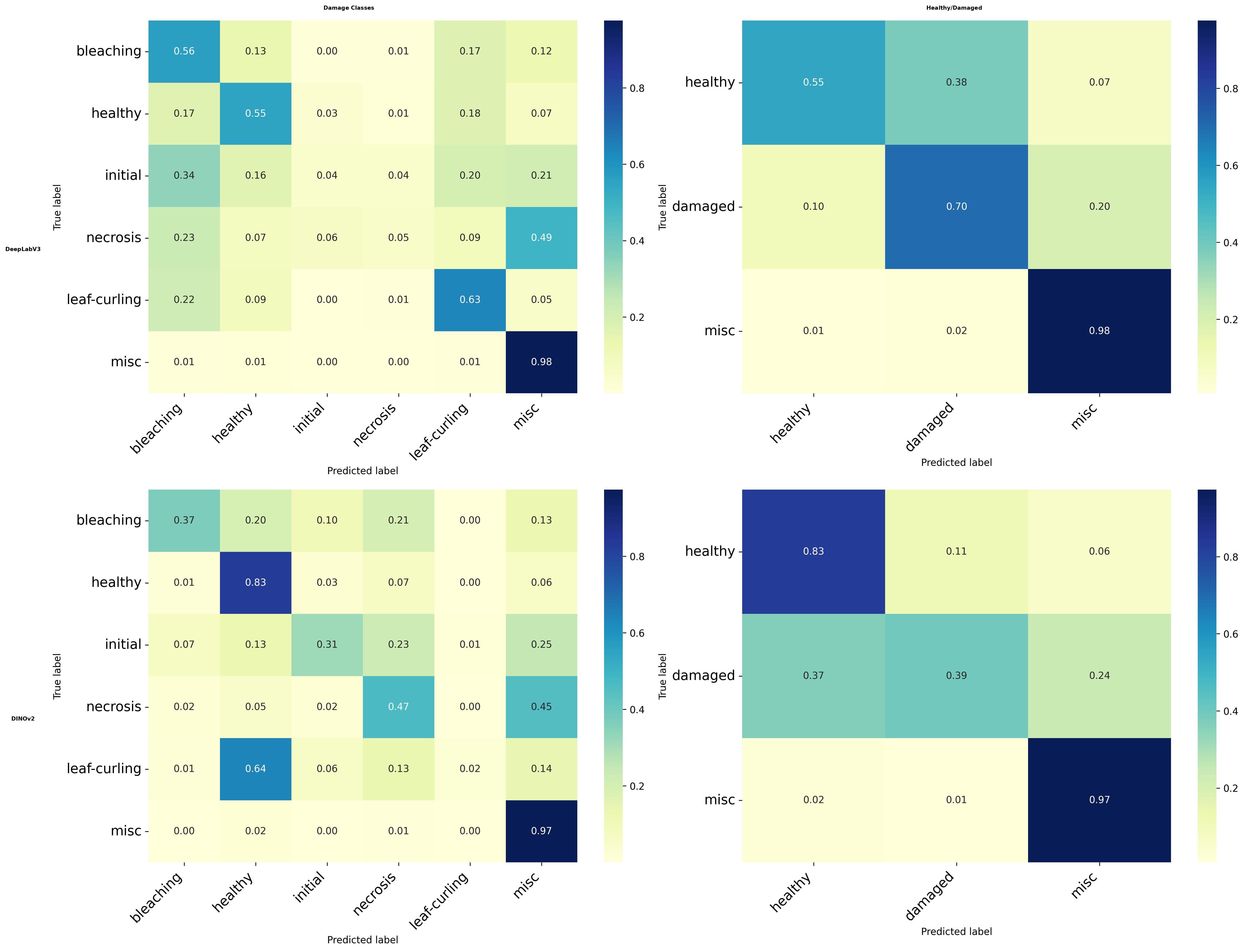}
  \caption{Damage identification confusion matrices for the DINOv2-based model (bottom) and the DeepLabV3-based model (top) on the REALITY dataset evaluated at different damage hierarchies: damage classes and healthy/damaged}
  \label{fig:reality_damage_camera_confusion_matrices}
\end{figure*}

Confusion matrices at different damage hierarchies can be seen in Figure~\ref{fig:reality_damage_camera_confusion_matrices}. It can be appreciated that damage classes cannot be properly distinguished among them. However, if we analyze the confusion matrix in more detail, most of the misclassifications occur between similar damage types. For example, initial damage is mostly confused with healthy vegetation, and necrosis is mostly confused with the misc class that covers soil and dead vegetation. In contrast, healthy vegetation is almost never confused with any other damage type or with the misc class. This is expected, as these damage types have similar visual characteristics, making them more challenging to differentiate among damage classes under domain shift conditions.

\begin{figure}[ht!]
  \centering
  \footnotesize
  \includegraphics[height=10cm]{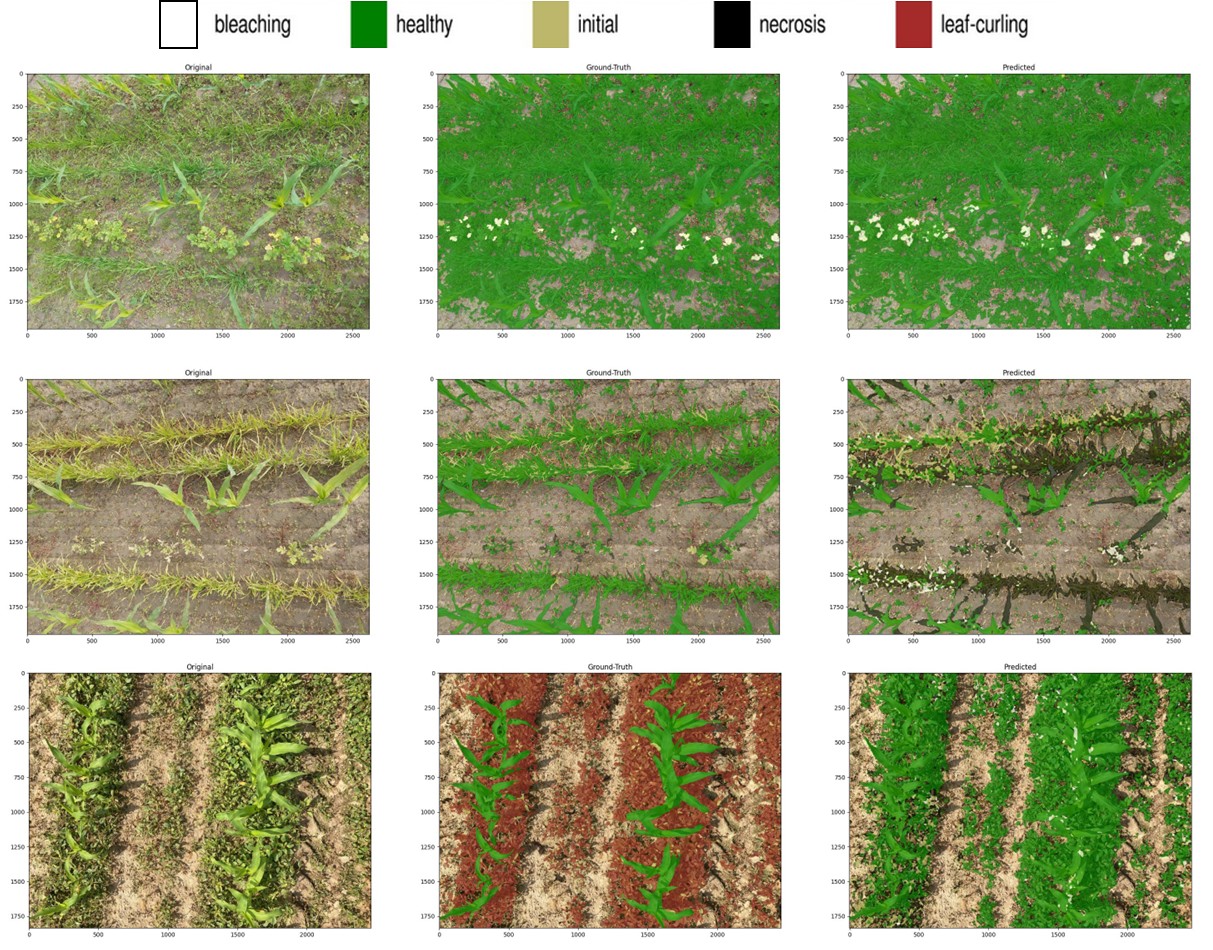}
  \caption{Algorithm damage predictions for \textit{Zea mays} field. Model trained on the BASE dataset and tested on the REALITY dataset}
  \label{fig:damage_results_reality_zeamx}
\end{figure}

The REALITY dataset contains a much higher percentage of damaged vegetation than the BASE dataset. This is particularly evident in the elevated prevalence of the misc class, which includes soil and dead vegetation with a very high level of damage. Another detected issue is the percentage of leaf-curling that was classified as healthy. As shown in Figure~\ref{fig:damage_results_reality_zeamx} for the \textit{Zea mays} crop, there is a high presence of leaf-curling in weed species. However, leaf-curling in weed species did not appear in the BASE dataset and was almost insignificant for crop species. Because of this, leaf-curling will be left out of the analysis on the confusion matrices, as it contaminates the performance analysis.

\begin{figure}[ht!]
  \centering
  \footnotesize
  \includegraphics[height=10cm]{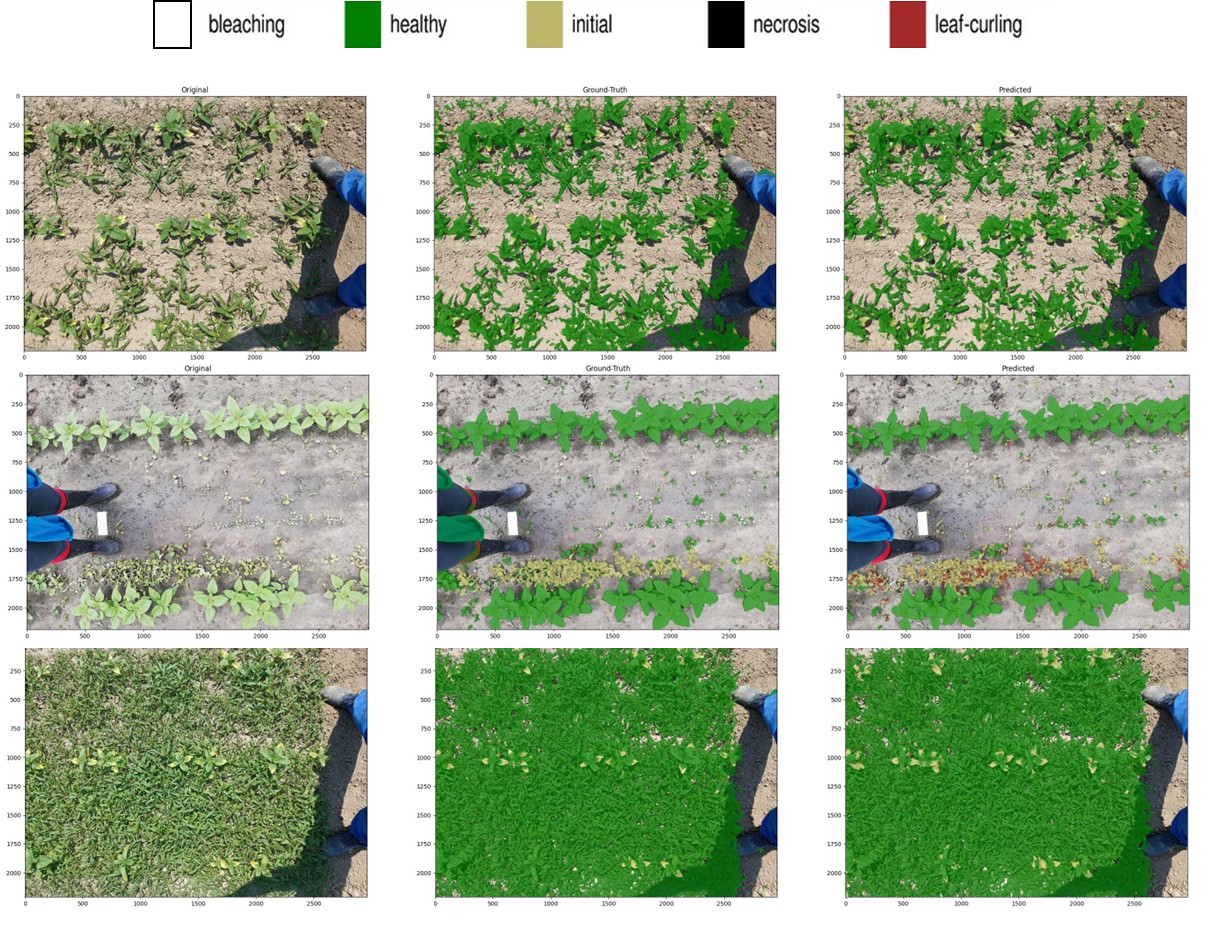}
  \caption{Algorithm damage predictions for \textit{Helianthus annuus} field. Model trained on the BASE dataset and tested on the REALITY dataset}
  \label{fig:damage_results_reality_helan}
\end{figure}

\begin{figure}[ht!]
  \centering
  \footnotesize
  \includegraphics[height=10cm]{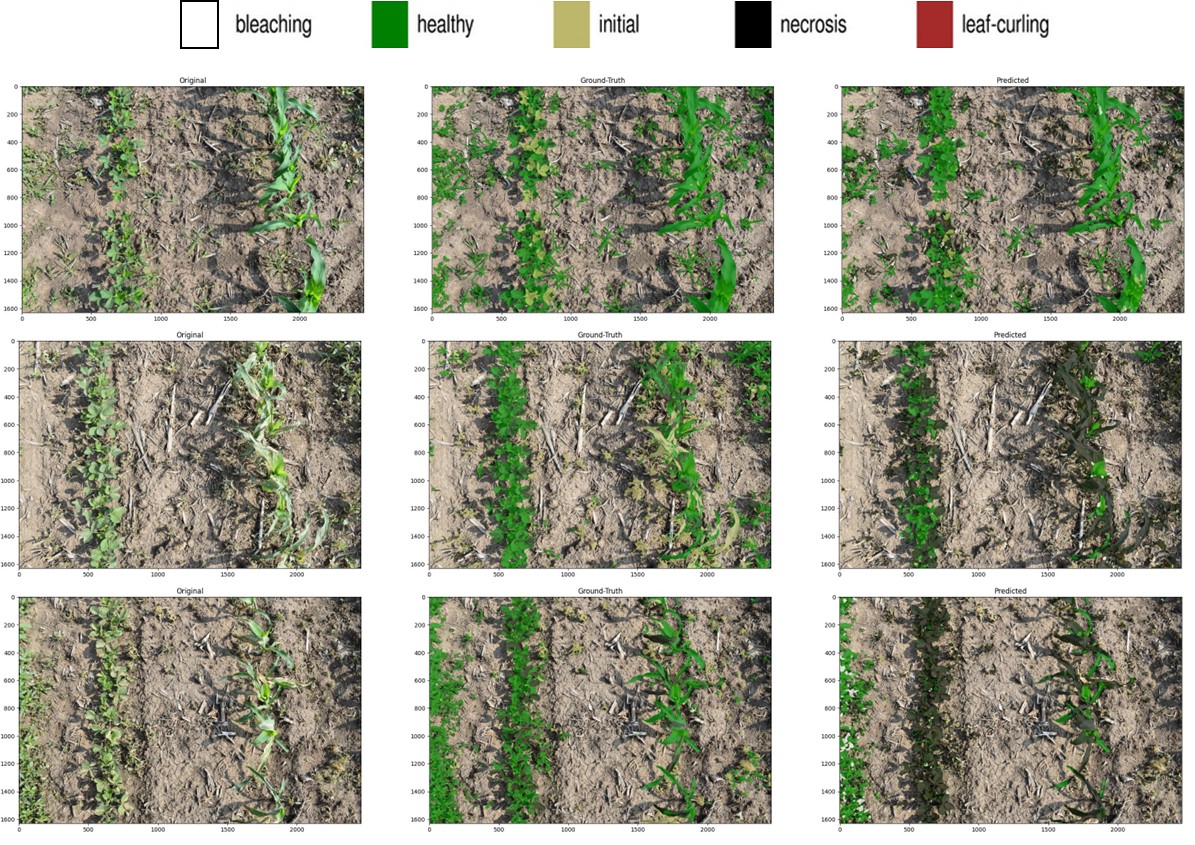}
  \caption{Algorithm damage predictions for \textit{Glycine max} field. Model trained on the BASE dataset and tested on the REALITY dataset}
  \label{fig:damage_results_reality_glxma}
\end{figure}

Algorithm predictions are illustrated in Figures~\ref{fig:damage_results_reality_helan}, \ref{fig:damage_results_reality_glxma}, and \ref{fig:damage_results_reality_zeamx} for \textit{Helianthus annuus}, \textit{Glycine max}, and \textit{Zea mays} fields, respectively. These visualizations demonstrate the model's capability to accurately identify various damage types within different crop environments.

Table~\ref{tab:reality_check_camera_damage_experiments_perclass_dinov2_damage_classes} provides a detailed breakdown of per-class performance metrics for damage identification using the DINOv2-based model on the REALITY dataset. The results indicate that damage type identification is more challenging under domain shift conditions. However, if we want to perform damage/healthy identification, including dead vegetation in the damaged class (misc class), the model performs quite well. Damage class identification remains challenging, with balanced accuracy values below 0.7 for most damage types except for healthy and misc classes.

\begin{table}[ht!]
  \centering
  \scriptsize
  \caption{Per-class damage identification performance metrics using the DINOv2-based model on the REALITY dataset.}
  \label{tab:reality_check_camera_damage_experiments_perclass_dinov2_damage_classes}
  \resizebox{0.95\textwidth}{!}{%
    \begin{tabular}{lccccc}
      \hline
      Class        & Balanced Accuracy & Precision (PPV)   & Recall (Sensitivity) & F1 Score          & R²                \\
      \hline
      bleaching    & $0.680 \pm 0.121$ & $0.022 \pm 0.022$ & $0.365 \pm 0.242$    & $0.042 \pm 0.039$ & $0.283 \pm 0.248$ \\
      healthy      & $0.893 \pm 0.021$ & $0.881 \pm 0.056$ & $0.825 \pm 0.040$    & $0.852 \pm 0.037$ & $0.767 \pm 0.164$ \\
      initial      & $0.650 \pm 0.085$ & $0.257 \pm 0.109$ & $0.308 \pm 0.171$    & $0.278 \pm 0.132$ & $0.466 \pm 0.370$ \\
      leaf-curling & $0.517 \pm 0.029$ & $0.336 \pm 0.303$ & $0.034 \pm 0.059$    & $0.051 \pm 0.094$ & $0.018 \pm 0.278$ \\
      misc         & $0.944 \pm 0.008$ & $0.963 \pm 0.009$ & $0.974 \pm 0.005$    & $0.969 \pm 0.007$ & $0.988 \pm 0.006$ \\
      necrosis     & $0.719 \pm 0.052$ & $0.152 \pm 0.039$ & $0.466 \pm 0.106$    & $0.228 \pm 0.052$ & $0.613 \pm 0.201$ \\
      \hline
    \end{tabular}%
  }
\end{table}



In summary, the proposed DINOv2-based model demonstrates improved performance in damage identification compared to the previous DeepLabV3-based model on the REALITY dataset, although both models face challenges under domain shift conditions. The detailed per-class analysis reveals that while healthy/damaged classification remains relatively robust, distinguishing between specific damage types is more challenging in unseen environments. These findings underscore the importance of developing models that can generalize across different environmental conditions and the value of self-supervised foundation models in achieving better domain robustness.

\subsection{Evaluation on the DRONE Dataset: Assessing Domain Shift Robustness from Different Sensor Modalities}
\label{ssec:results_drone}

Evaluations were conducted on the DRONE dataset to assess the robustness of the proposed DINOv2-based model under domain shift conditions arising from different sensor modalities not seen during training (drone imagery). The performance of the DINOv2-based model was compared against the previous DeepLabV3-based model across various taxonomical hierarchies for species identification. Table~\ref{tab:reality_check_drone_species_experiments_comparison} presents the results of this comparison. It is evident that the DINOv2-based model significantly outperforms the DeepLabV3-based model across all taxonomical levels, demonstrating its superior ability to generalize to data from different sensor modalities.

\begin{table}[ht!]
  \centering
  \scriptsize
  \caption{Comparison of species identification performance between DeepLabV3 and DINOv2 models across different taxonomical hierarchies on the DRONE dataset.}
  \label{tab:reality_check_drone_species_experiments_comparison}
  \resizebox{0.95\textwidth}{!}{%
    \begin{tabular}{llccccc}
      \hline
      Experiment & Hierarchy  & Balanced Accuracy          & Precision (PPV)            & Recall (Sensitivity)       & F1 Score                   & R²                         \\
      \hline
      DeepLabV3  & Species    & $0.576 \pm 0.030$          & $0.237 \pm 0.054$          & $0.166 \pm 0.059$          & $0.139 \pm 0.032$          & $0.163 \pm 0.103$          \\
      DINOv2     & Species    & \textbf{$0.791 \pm 0.049$} & \textbf{$0.444 \pm 0.076$} & \textbf{$0.591 \pm 0.097$} & \textbf{$0.444 \pm 0.080$} & \textbf{$0.498 \pm 0.146$} \\
      DeepLabV3  & Genus      & $0.588 \pm 0.030$          & $0.252 \pm 0.052$          & $0.192 \pm 0.059$          & $0.168 \pm 0.042$          & $0.186 \pm 0.109$          \\
      DINOv2     & Genus      & \textbf{$0.796 \pm 0.042$} & \textbf{$0.467 \pm 0.085$} & \textbf{$0.600 \pm 0.085$} & \textbf{$0.482 \pm 0.081$} & \textbf{$0.542 \pm 0.133$} \\
      DeepLabV3  & Family     & $0.631 \pm 0.033$          & $0.352 \pm 0.058$          & $0.288 \pm 0.065$          & $0.272 \pm 0.047$          & $0.250 \pm 0.098$          \\
      DINOv2     & Family     & \textbf{$0.877 \pm 0.033$} & \textbf{$0.660 \pm 0.097$} & \textbf{$0.768 \pm 0.065$} & \textbf{$0.681 \pm 0.085$} & \textbf{$0.782 \pm 0.155$} \\
      DeepLabV3  & Order      & $0.670 \pm 0.039$          & $0.438 \pm 0.066$          & $0.372 \pm 0.076$          & $0.357 \pm 0.061$          & $0.366 \pm 0.151$          \\
      DINOv2     & Order      & \textbf{$0.904 \pm 0.020$} & \textbf{$0.748 \pm 0.071$} & \textbf{$0.826 \pm 0.040$} & \textbf{$0.764 \pm 0.061$} & \textbf{$0.802 \pm 0.114$} \\
      DeepLabV3  & Class      & $0.838 \pm 0.023$          & $0.827 \pm 0.060$          & $0.740 \pm 0.040$          & $0.752 \pm 0.052$          & $0.707 \pm 0.209$          \\
      DINOv2     & Class      & \textbf{$0.935 \pm 0.009$} & \textbf{$0.857 \pm 0.061$} & \textbf{$0.913 \pm 0.014$} & \textbf{$0.880 \pm 0.040$} & \textbf{$0.728 \pm 0.198$} \\
      DeepLabV3  & Vegetation & $0.911 \pm 0.010$          & \textbf{$0.933 \pm 0.033$} & $0.911 \pm 0.013$          & \textbf{$0.921 \pm 0.017$} & \textbf{$0.831 \pm 0.151$} \\
      DINOv2     & Vegetation & \textbf{$0.935 \pm 0.008$} & $0.883 \pm 0.043$          & \textbf{$0.935 \pm 0.013$} & $0.907 \pm 0.028$          & $0.636 \pm 0.241$          \\
      \hline
    \end{tabular}%
  }
\end{table}

Figure~\ref{fig:reality_species_drone_confusion_matrices} shows the confusion matrices for species identification at different taxonomical hierarchies using both the DINOv2-based model and the DeepLabV3-based model on the DRONE dataset. The DINOv2 model exhibits higher accuracy and fewer misclassifications across all hierarchical levels compared to the DeepLabV3 model.

\begin{figure*}[ht!]
  \centering
  \footnotesize
  \includegraphics[width=14cm]{reality_check_drone_confusion_matrices.jpg}
  \caption{Species identification confusion matrices for the DINOv2-based model (bottom) and the DeepLabV3-based model (top) on the DRONE dataset evaluated at different taxonomical hierarchies: species, genus, family, order, class}
  \label{fig:reality_species_drone_confusion_matrices}
\end{figure*}

\begin{figure}[ht!]
  \centering
  \footnotesize
  \includegraphics[height=10cm]{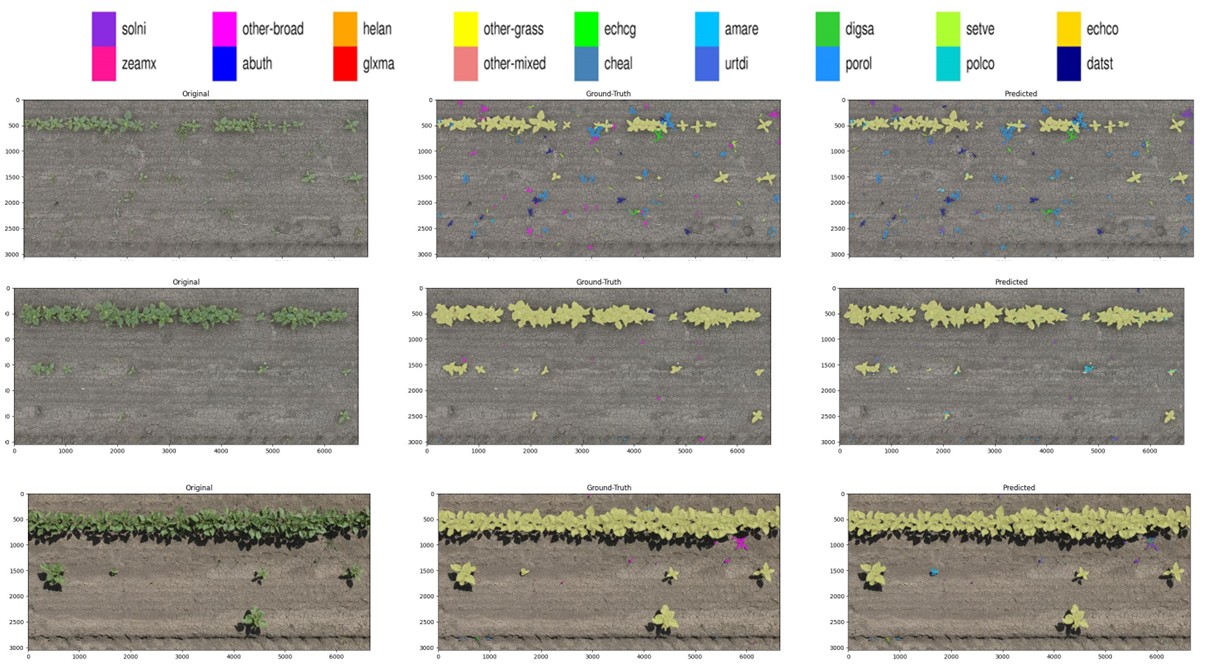}
  \caption{Algorithm species predictions for \textit{Helianthus annuus} field. Model trained on the BASE dataset and tested on the DRONE dataset}
  \label{fig:species_results_reality_helan_drone}
\end{figure}

\begin{figure}[ht!]
  \centering
  \footnotesize
  \includegraphics[height=10cm]{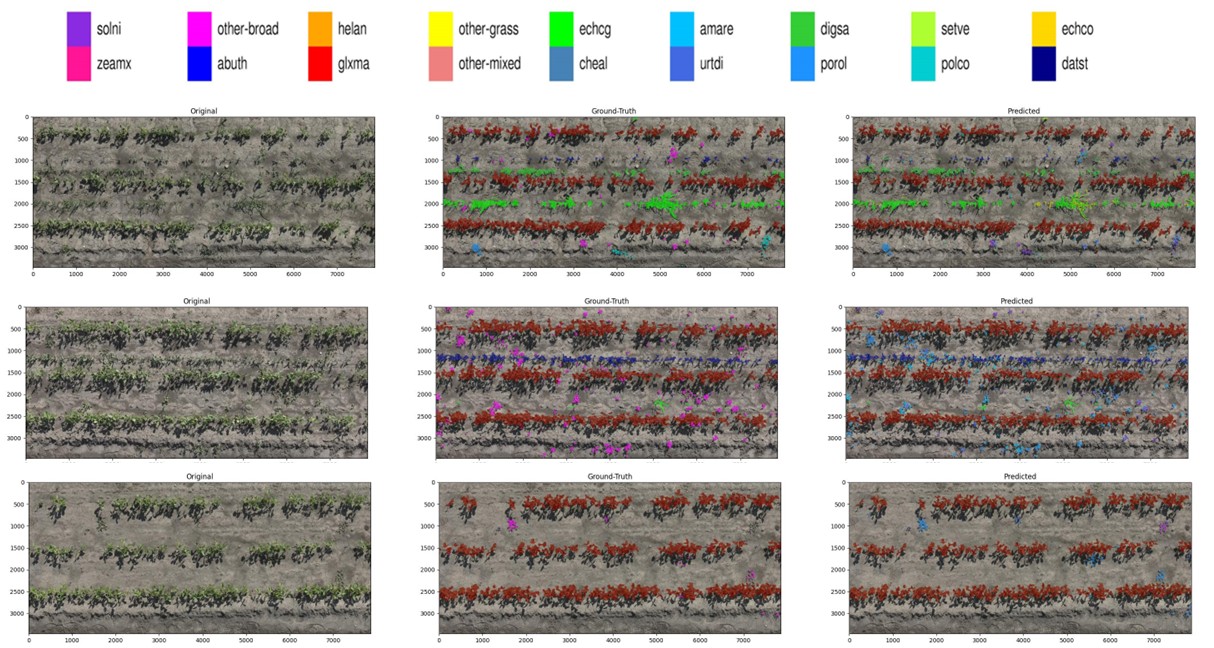}
  \caption{Algorithm species predictions for \textit{Glycine max} field. Model trained on the BASE dataset and tested on the DRONE dataset}
  \label{fig:species_results_reality_glxma_drone}
\end{figure}

\begin{figure}[ht!]
  \centering
  \footnotesize
  \includegraphics[height=10cm]{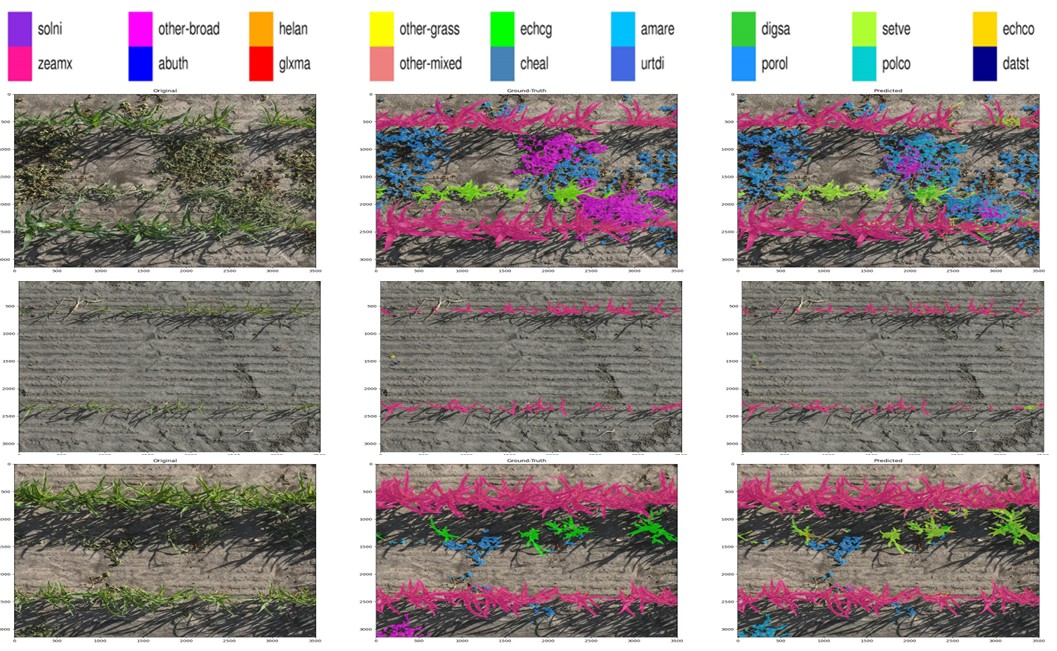}
  \caption{Algorithm species predictions for \textit{Zea mays} field. Model trained on the BASE dataset and tested on the DRONE dataset}
  \label{fig:species_results_reality_zeamx_drone}
\end{figure}

Results of algorithm predictions are illustrated in Figures~\ref{fig:species_results_reality_helan_drone}, \ref{fig:species_results_reality_glxma_drone}, and \ref{fig:species_results_reality_zeamx_drone} for \textit{Helianthus annuus}, \textit{Glycine max}, and \textit{Zea mays} fields, respectively. These visualizations demonstrate the model's capability to accurately identify various species within different crop environments using drone imagery.

If we analyze the per-class results in Table~\ref{tab:reality_check_drone_species_experiments_perclass_dinov2_species}, we can see that the DINOv2-based model performs well for several species, achieving balanced accuracy values above 0.8 for species such as \textit{Abutilon theophrasti}, \textit{Amaranthus retroflexus}, \textit{Datura stramonium}, \textit{Glycine max}, \textit{Helianthus annuus}, \textit{Portulaca oleracea}, and \textit{Zea mays}. However, certain species such as \textit{Chenopodium album}, \textit{Echinochloa colona}, \textit{Setaria verticillata}, and \textit{Solanum nigrum} exhibit relatively lower performance, with balanced accuracy values below 0.7. These results highlight the challenges associated with species identification under domain shift conditions, particularly when dealing with data from different sensor modalities. However, most misclassifications occur among species and the misc class that arises from damaged vegetation and soil, which is expected given the elevated presence of damaged vegetation in the DRONE dataset.

\begin{table}[ht!]
  \centering
  \scriptsize
  \caption{Per-class performance metrics for species identification using the DINOv2-based model on the DRONE dataset.}
  \label{tab:reality_check_drone_species_experiments_perclass_dinov2_species}
  \resizebox{0.95\textwidth}{!}{%
    \begin{tabular}{lccccc}
      \hline
      Class & Balanced Accuracy & Precision (PPV)   & Recall (Sensitivity) & F1 Score          & R²                \\
      \hline
      abuth & $0.835 \pm 0.034$ & $0.629 \pm 0.134$ & $0.671 \pm 0.067$    & $0.648 \pm 0.096$ & $0.846 \pm 0.128$ \\
      amare & $0.867 \pm 0.031$ & $0.095 \pm 0.082$ & $0.749 \pm 0.064$    & $0.166 \pm 0.124$ & $0.081 \pm 0.193$ \\
      cheal & $0.631 \pm 0.103$ & $0.061 \pm 0.054$ & $0.264 \pm 0.205$    & $0.093 \pm 0.066$ & $0.060 \pm 0.144$ \\
      datst & $0.866 \pm 0.038$ & $0.665 \pm 0.147$ & $0.732 \pm 0.075$    & $0.695 \pm 0.100$ & $0.957 \pm 0.057$ \\
      digsa & $0.814 \pm 0.088$ & $0.104 \pm 0.100$ & $0.630 \pm 0.176$    & $0.175 \pm 0.150$ & $0.071 \pm 0.154$ \\
      echcg & $0.691 \pm 0.092$ & $0.746 \pm 0.131$ & $0.383 \pm 0.184$    & $0.498 \pm 0.154$ & $0.695 \pm 0.274$ \\
      echco & $0.845 \pm 0.097$ & $0.001 \pm 0.001$ & $0.694 \pm 0.194$    & $0.002 \pm 0.002$ & $0.078 \pm 0.345$ \\
      glxma & $0.949 \pm 0.013$ & $0.931 \pm 0.011$ & $0.899 \pm 0.027$    & $0.915 \pm 0.012$ & $0.983 \pm 0.011$ \\
      helan & $0.948 \pm 0.017$ & $0.969 \pm 0.006$ & $0.896 \pm 0.034$    & $0.931 \pm 0.019$ & $0.990 \pm 0.011$ \\
      misc  & $0.935 \pm 0.008$ & $0.990 \pm 0.002$ & $0.974 \pm 0.012$    & $0.982 \pm 0.006$ & $0.636 \pm 0.248$ \\
      panmi & $0.500 \pm 0.000$ & $0.000 \pm 0.000$ & $0.000 \pm 0.000$    & $0.000 \pm 0.000$ & $0.000 \pm 0.000$ \\
      polco & $0.717 \pm 0.107$ & $0.326 \pm 0.241$ & $0.435 \pm 0.215$    & $0.365 \pm 0.220$ & $0.605 \pm 0.440$ \\
      porol & $0.865 \pm 0.046$ & $0.523 \pm 0.204$ & $0.734 \pm 0.092$    & $0.606 \pm 0.167$ & $0.937 \pm 0.147$ \\
      setve & $0.774 \pm 0.058$ & $0.081 \pm 0.066$ & $0.552 \pm 0.117$    & $0.139 \pm 0.100$ & $0.067 \pm 0.167$ \\
      solni & $0.542 \pm 0.027$ & $0.037 \pm 0.026$ & $0.085 \pm 0.055$    & $0.050 \pm 0.031$ & $0.001 \pm 0.003$ \\
      zeamx & $0.879 \pm 0.020$ & $0.953 \pm 0.015$ & $0.758 \pm 0.041$    & $0.844 \pm 0.026$ & $0.967 \pm 0.016$ \\
      \hline
    \end{tabular}%
  }
\end{table}


When analyzing the performance per location, Table~\ref{tab:reality_check_drone_species_experiments_perlocation_dinov2_species} presents the results for Germany (DE), Spain (ES), and the United States (US). The DINOv2-based model consistently outperforms the DeepLabV3-based model across all locations. Notably, the model achieves the highest balanced accuracy in Spain ($0.775 \pm 0.059$) and the lowest in the United States ($0.742 \pm 0.068$). These results highlight the model's robustness and adaptability to different geographical regions, although performance may vary based on the similarity of the test data to the training data.

\begin{table}[ht!]
  \centering
  \scriptsize
  \caption{Per-location species identification performance metrics using the DINOv2-based model on the DRONE dataset.}
  \label{tab:reality_check_drone_species_experiments_perlocation_dinov2_species}
  \resizebox{0.95\textwidth}{!}{%
    \begin{tabular}{llccccc}
      \hline
      Experiment & Location & Balanced Accuracy & Precision (PPV)   & Recall (Sensitivity) & F1 Score          & R²                \\
      \hline
      DINOv2     & DE       & $0.774 \pm 0.090$ & $0.346 \pm 0.084$ & $0.557 \pm 0.180$    & $0.324 \pm 0.081$ & $0.476 \pm 0.222$ \\
      DINOv2     & ES       & $0.775 \pm 0.059$ & $0.516 \pm 0.065$ & $0.560 \pm 0.117$    & $0.495 \pm 0.072$ & $0.580 \pm 0.130$ \\
      DINOv2     & US       & $0.742 \pm 0.068$ & $0.232 \pm 0.051$ & $0.493 \pm 0.137$    & $0.247 \pm 0.068$ & $0.235 \pm 0.217$ \\
      DeepLabV3  & DE       & $0.592 \pm 0.054$ & $0.266 \pm 0.033$ & $0.197 \pm 0.104$    & $0.109 \pm 0.028$ & $0.254 \pm 0.212$ \\
      DeepLabV3  & ES       & $0.557 \pm 0.028$ & $0.312 \pm 0.054$ & $0.132 \pm 0.054$    & $0.112 \pm 0.038$ & $0.209 \pm 0.165$ \\
      DeepLabV3  & US       & $0.564 \pm 0.029$ & $0.205 \pm 0.019$ & $0.145 \pm 0.056$    & $0.130 \pm 0.021$ & $0.148 \pm 0.111$ \\
      \hline
      DINOv2     & All      & $0.791 \pm 0.049$ & $0.444 \pm 0.076$ & $0.591 \pm 0.097$    & $0.444 \pm 0.080$ & $0.498 \pm 0.146$ \\
      DeepLabV3  & All      & $0.576 \pm 0.030$ & $0.237 \pm 0.054$ & $0.166 \pm 0.059$    & $0.139 \pm 0.032$ & $0.163 \pm 0.103$ \\
      \hline
    \end{tabular}%
  }
\end{table}

Most of the misclassifications occur between the species and the misc class, which includes damaged vegetation and soil. This is expected, as the REALITY dataset contains a high percentage of damaged vegetation, which is not present in the BASE dataset used for training. To better analyze the misclassifications between species, we present the data without the misc class below.

\begin{table}[ht!]
  \centering
  \scriptsize
  \caption{Comparison of species identification performance between DeepLabV3 and DINOv2 models across different taxonomical hierarchies on the DRONE dataset excluding the misc class.}
  \label{tab:reality_check_drone_species_experiments_comparison_nomisc}
  \resizebox{0.95\textwidth}{!}{%
    \begin{tabular}{llccccc}
      \hline
      Experiment & Hierarchy & Balanced Accuracy          & Precision (PPV)            & Recall (Sensitivity)       & F1 Score                   & R²                         \\
      \hline
      DeepLabV3  & Species   & $0.546 \pm 0.043$          & $0.197 \pm 0.058$          & $0.147 \pm 0.084$          & $0.095 \pm 0.040$          & $0.053 \pm 0.048$          \\
      DINOv2     & Species   & \textbf{$0.819 \pm 0.057$} & \textbf{$0.532 \pm 0.080$} & \textbf{$0.647 \pm 0.114$} & \textbf{$0.527 \pm 0.087$} & \textbf{$0.432 \pm 0.152$} \\
      DeepLabV3  & Genus     & $0.560 \pm 0.042$          & $0.213 \pm 0.058$          & $0.176 \pm 0.080$          & $0.128 \pm 0.053$          & $0.054 \pm 0.047$          \\
      DINOv2     & Genus     & \textbf{$0.826 \pm 0.049$} & \textbf{$0.572 \pm 0.085$} & \textbf{$0.659 \pm 0.099$} & \textbf{$0.580 \pm 0.086$} & \textbf{$0.488 \pm 0.140$} \\
      DeepLabV3  & Family    & $0.586 \pm 0.046$          & $0.299 \pm 0.065$          & $0.254 \pm 0.088$          & $0.213 \pm 0.059$          & $0.114 \pm 0.054$          \\
      DINOv2     & Family    & \textbf{$0.922 \pm 0.036$} & \textbf{$0.820 \pm 0.091$} & \textbf{$0.849 \pm 0.073$} & \textbf{$0.825 \pm 0.083$} & \textbf{$0.677 \pm 0.164$} \\
      DeepLabV3  & Order     & $0.623 \pm 0.054$          & $0.377 \pm 0.073$          & $0.351 \pm 0.106$          & $0.291 \pm 0.077$          & $0.151 \pm 0.062$          \\
      DINOv2     & Order     & \textbf{$0.954 \pm 0.018$} & \textbf{$0.900 \pm 0.049$} & \textbf{$0.915 \pm 0.036$} & \textbf{$0.905 \pm 0.040$} & \textbf{$0.824 \pm 0.121$} \\
      DeepLabV3  & Class     & $0.733 \pm 0.054$          & $0.839 \pm 0.064$          & $0.734 \pm 0.062$          & $0.744 \pm 0.076$          & $0.464 \pm 0.125$          \\
      DINOv2     & Class     & \textbf{$0.991 \pm 0.002$} & \textbf{$0.991 \pm 0.004$} & \textbf{$0.991 \pm 0.003$} & \textbf{$0.991 \pm 0.003$} & \textbf{$0.995 \pm 0.005$} \\

      \hline
    \end{tabular}%
  }
\end{table}

Species identification confusion matrices excluding the misc class at different taxonomical hierarchies can be seen in Figure~\ref{fig:reality_species_drone_confusion_matrices_nomisc}. It can be appreciated that the DINOv2-based model still outperforms the DeepLabV3-based model significantly, achieving higher accuracy and fewer misclassifications across all hierarchical levels and obtaining clearer distinctions between species.

\begin{figure*}[ht!]
  \centering
  \footnotesize
  \includegraphics[width=14cm]{reality_check_drone_confusion_matrices_nomisc.jpg}
  \caption{Species identification confusion matrices for the DINOv2-based model (bottom) and the DeepLabV3-based model (top) on the DRONE dataset evaluated at different taxonomical hierarchies: species, genus, family, order, class excluding the misc class.}
  \label{fig:reality_species_drone_confusion_matrices_nomisc}
\end{figure*}

In summary, the DINOv2-based model demonstrates significantly improved performance compared to the DeepLabV3-based model when evaluated on the DRONE dataset, despite the substantial domain shift introduced by the different sensor modality. While performance is lower than on the BASE dataset, the DINOv2 model shows remarkable robustness in generalizing from mobile phone and digital camera images to drone imagery. The results confirm that self-supervised foundation models like DINOv2 provide better transferability across different imaging platforms, making them particularly valuable for agricultural applications where multiple sensor types may be deployed. The higher performance on Spanish locations compared to the U.S. locations indicates that geographical similarity to the training data remains an important factor, suggesting potential benefits from incorporating more diverse geographical data in future training datasets.

\section{Implementation and deployment}

The models were implemented using PyTorch Lightning and trained on Tecnalia's KATEA computing cluster equipped with Nvidia H100 nodes. The resulting models were deployed in the BASF system for biological trial management and assessment.



\section{Discussion}

The DINOv2-based model incorporating hierarchical inference demonstrates marked improvements over the DeepLabV3+ baseline across all evaluated datasets. These results indicate robust generalization capabilities under a range of domain shift scenarios.

\subsection{Model Performance Across Domain Shifts}

\subsubsection{Baseline Performance}
When trained and evaluated on the BASE dataset, the proposed model achieved substantial improvements relative to the baseline. The species identification F1 score increased from $0.53 \pm 0.06$ (DeepLabV3) to $0.87 \pm 0.03$ (DINOv2), corresponding to a 63\% relative improvement. Similarly, the damage classification F1 score improved from $0.38 \pm 0.07$ to $0.64 \pm 0.05$, a 69\% relative increase. The high $R^2$ values ($0.98 \pm 0.02$ for species, $0.87 \pm 0.07$ for damage) further support the reliability of the quantitative predictions. Most species achieved balanced accuracies exceeding 0.9; however, certain species (e.g., \textit{Echinochloa colona}, \textit{Solanum nigrum}) exhibited lower performance, likely attributable to visual similarity and limited training data. For damage classification, the detection of subtle early-stage symptoms remains a notable challenge.

\subsubsection{Moderate Domain Shift: REALITY Dataset}
The REALITY dataset (2023 images from Germany, Spain, and the United States) was used to assess generalization across years, regions, and environmental conditions. The DINOv2 model maintained robust performance, with a species F1 score of $0.77 \pm 0.10$ (a reduction of only 0.10 from the BASE dataset), whereas the DeepLabV3 model experienced a more pronounced decline (from $0.53 \pm 0.06$ to $0.24 \pm 0.08$), corresponding to a 215\% relative improvement for DINOv2 under domain shift. The $R^2$ value also remained high for DINOv2 ($0.88 \pm 0.13$) compared to the baseline ($0.19 \pm 0.16$), indicating superior quantitative reliability.

Geographic variation was observed: Spain achieved the highest accuracy ($0.91 \pm 0.05$), followed by Germany ($0.83 \pm 0.06$), and the United States ($0.73 \pm 0.08$). This pattern reflects the distribution of training data (Germany and Spain included, United States excluded). Notably, even in the previously unseen United States location, the DINOv2 model substantially outperformed the baseline (F1: $0.43 \pm 0.14$ vs. $0.17 \pm 0.05$).

The REALITY dataset introduced novel damage patterns (e.g., leaf-curling in weeds) absent from the training data, resulting in increased confusion among damage classes. Nevertheless, the model maintained satisfactory discrimination between healthy and damaged plants (F1: $0.78 \pm 0.03$), which is often the primary practical requirement.

\subsubsection{Extreme Domain Shift: DRONE Dataset}
The DRONE dataset constitutes an extreme domain shift, involving a transition from ground-level to aerial imagery at an altitude of 6 meters. Although species-level performance declined (F1: $0.44 \pm 0.08$), this result should be interpreted in the context of the DeepLabV3 baseline's near-complete failure (F1: $0.14 \pm 0.03$), representing a 214\% relative improvement. The hierarchical inference mechanism was particularly beneficial in this setting, preserving strong performance at higher taxonomic levels: genus (F1: $0.48 \pm 0.08$), family (F1: $0.68 \pm 0.09$), order (F1: $0.76 \pm 0.06$), and class (F1: $0.88 \pm 0.04$).

When the miscellaneous class (dead vegetation and soil) is excluded, performance improves further at the species (F1: $0.53 \pm 0.09$), genus (F1: $0.58 \pm 0.09$), and family (F1: $0.83 \pm 0.08$) levels. This finding suggests that most errors are attributable to confusion between species and soil, rather than misclassification among species. Thus, the primary challenge lies in distinguishing vegetation from severely damaged or soil-like backgrounds.

Such hierarchical resilience is essential for practical deployment, as it enables reliable coarse-grained classifications (e.g., crop vs. weed, broadleaf vs. grass) when fine-grained species identification is uncertain. The system can adaptively report broader classifications based on confidence thresholds, thereby enhancing robustness in challenging scenarios.

\subsection{Factors Contributing to Improved Performance}

The observed performance gains are attributable to several key design choices. First, DINOv2's self-supervised pre-training on large-scale, diverse datasets yields robust visual features that transfer effectively to agricultural domains, capturing patterns that generalize more successfully than those learned via supervised ImageNet pre-training. Second, hierarchical inference mitigates the dispersion of probability mass across visually similar species and enables graceful degradation under uncertainty. This property is particularly valuable when domain shift reduces species-level reliability but preserves higher-level accuracy. Finally, the BASE dataset's inclusion of multiple years (2018--2020), locations (Germany, Spain), and devices provides sufficient variability for learning robust features, although further geographic expansion would likely enhance generalization.

The model's robustness to temporal, geographic, device, and sensor shifts renders it suitable for real-world herbicide trial evaluation. It is currently deployed in BASF's phenotyping pipeline for automated crop and weed monitoring. The hierarchical output structure enables flexible reporting, ranging from fine-grained species identification to broad vegetation coverage, depending on confidence levels and application requirements. The model's ability to generate meaningful predictions under extreme domain shift (e.g., ground-to-aerial transfer) underscores the potential of foundation model approaches in agricultural computer vision.

\subsection{Limitations and Future Directions}

Despite the substantial improvements observed, several limitations merit consideration. The geographic scope of the training data is restricted to Germany and Spain (2018--2020), which contributes to reduced performance on United States locations. The gap in sensor modalities is also evident in the drone imagery results (species F1: $0.44 \pm 0.08$), although most errors are attributable to species-soil confusion rather than inter-species misclassification, potentially reflecting annotation inconsistencies. While this study demonstrates the utility of foundation models under severe domain drift, production deployments should incorporate training data from all relevant geographies and sensor modalities to optimize performance.

Damage classification remains a challenging task, particularly for subtle damage types (e.g., initial damage F1: $0.28 \pm 0.13$) and patterns absent from the training data (e.g., leaf-curling in weeds: F1: $0.05 \pm 0.09$). Evolving annotation protocols and subjective criteria further complicate this task. Additionally, limited representation of certain species (e.g., \textit{Echinochloa colona}, \textit{Solanum nigrum}) results in reduced performance, despite the use of effective sample weighting.

In terms of computational requirements, the use of a frozen DINOv2 backbone with only the multi-task decoder trained ensures computational efficiency comparable to EfficientNet-based approaches.

Although the model has been validated for herbicide field trials, further validation is necessary for other agricultural applications.

Furthermore, the absence of explicit uncertainty quantification and explainability currently limits the automated flagging of uncertain predictions for human review. The development of such capabilities is planned for future work.

Future research should prioritize expanding the diversity of training data across geographic regions and sensor modalities, developing methods for uncertainty quantification, and refining damage annotation protocols to address subjective criteria and capture emerging damage patterns.

\section{Conclusions}
\label{sec:conclusions}

This study evaluates the effect of domain drift on multi‑species plant segmentation and herbicide‑damage identification, and quantifies how self‑supervised vision foundation models (DINOv2) change robustness in realistic agricultural scenarios. Rather than proposing a novel segmentation algorithm, our emphasis is empirical: we measure performance degradation across seasons, geographies, devices, and sensor modalities, and test whether large pretrained visual representations reduce those degradations in operational phenotyping contexts. Experiments span multiple years and platforms (smartphones, handheld cameras, and drones) and include herbicide field trials covering more than 70 species, exposing models to substantial variability in acquisition conditions.

The three‑stage evaluation delivers three core findings. First, models using a DINOv2 backbone substantially outperform a DeepLabV3+ baseline on in‑distribution data (species F1: 0.87 vs. 0.52) and retain more performance under realistic shifts (moderate temporal/device shift: 0.77 vs. 0.24; extreme aerial shift: 0.44 vs. 0.14). Second, hierarchical taxonomic inference provides useful robustness: when fine‑grained species predictions degrade (particularly on aerial imagery), coarser taxonomic outputs (family F1: 0.68, class F1: 0.88) remain informative for downstream phenotyping tasks. Third, remaining failures under the most severe shifts are dominated by vegetation–soil separation and viewpoint changes rather than wholesale loss of species‑specific cues, suggesting pretrained features preserve relevant biological distinctions even as background and scale vary.

Deployment within BASF’s phenotyping pipeline offers practical confirmation that foundation‑backboned models can improve operational performance compared to conventional baselines. That said, the training distribution remains concentrated in Germany and Spain (2018–2020), which likely contributes to reduced absolute performance at U.S. sites despite consistent relative gains versus the baseline.

Limitations point to clear, empirical next steps. Damage classification for early or rare symptom patterns remains challenging (e.g., initial damage F1: 0.28; uncommon damage types F1: 0.05), and drone imagery shows persistent vegetation–soil confusion; both issues likely reflect annotation coverage and representational gaps for aerial viewpoints. Incorporating more aerial and geographically diverse training data, standardizing damage annotations, and adding uncertainty estimation would directly address observed failure modes and improve automated quality control.

In summary, our systematic cross‑domain evaluation demonstrates that self‑supervised foundation visual models materially mitigate the effects of domain drift in complex agricultural settings. The quantitative results and error analyses here are intended as a practical guide for deploying robust segmentation systems in herbicide research trials: they show where pretrained representations help most, where failures persist, and which dataset expansions or controls are likely to yield the greatest operational benefit.

\section{Data Availability Statement}
\label{sec:data_availability}
The data analyzed in this study is subject to the following licenses/restrictions. The dataset used in this article has been generated by the BASF R\&D field research community. It could be made available on reasonable request for non-commercial research purposes and under an agreement with BASF. Requests to access these datasets should be directed to ramon.navarra-mestre@basf.com.

\FloatBarrier

\bibliography{mybibfile}

\end{document}